\documentclass[journal]{IEEEtran}
\usepackage{amsmath,amsfonts}
\usepackage{array}
\usepackage[caption=false,font=normalsize,labelfont=sf,textfont=sf]{subfig}
\usepackage{textcomp}
\usepackage{stfloats}
\usepackage{url}
\usepackage{verbatim}
\usepackage{graphicx}
\usepackage{cite}
\usepackage{booktabs}
\usepackage{xcolor}
\usepackage{array,multirow}
\usepackage{float}
\usepackage{graphicx}
\usepackage{subfig}
\usepackage{amsmath}
\usepackage{wrapfig}
\usepackage{pifont}
\newcommand{\cmark}{\ding{51}}
\newcommand{\xmark}{\ding{55}}

\usepackage[ruled,vlined]{algorithm2e}
\usepackage{setspace}
\usepackage{etoolbox}
\AtBeginEnvironment{algorithm}{\SetArgSty{textrm}}
\SetKwFor{For}{for (}{) $\lbrace$}{$\rbrace$}

\usepackage[colorlinks]{hyperref}

\begin{document}

\title{Dynamic Frame Interpolation in Wavelet Domain}

\author{Lingtong Kong, Boyuan Jiang, Donghao Luo, Wenqing Chu, Ying Tai, Chengjie Wang, Jie Yang
	\thanks{Lingtong Kong and Jie Yang are with the Institute of Image Processing and Pattern Recognition, Department of Automation, Shanghai Jiao Tong University, Shanghai 200240, China (e-mail: ltkong218@gmail.com, jieyang@sjtu.edu.cn).}
	\thanks{Boyuan Jiang, Donghao Luo, Wenqing Chu and Chengjie Wang are with the Youtu Lab, Tencent, Shanghai 200233, China (e-mail: byronjiang@tencent.com, michaelluo@tencent.com, wqchu16@gmail.com, jasoncjwang@tencent.com).}
	\thanks{Ying Tai is with the School of Intelligence Science and Technology, Nanjing University, Suzhou 215163, China (e-mail: yingtai@nju.edu.cn).}
}



\maketitle

\begin{abstract}
Video frame interpolation is an important low-level vision task, which can increase frame rate for more fluent visual experience. Existing methods have achieved great success by employing advanced motion models and synthesis networks. However, the spatial redundancy when synthesizing the target frame has not been fully explored, that can result in lots of inefficient computation. On the other hand, the computation compression degree in frame interpolation is highly dependent on both texture distribution and scene motion, which demands to understand the spatial-temporal information of each input frame pair for a better compression degree selection. In this work, we propose a novel two-stage frame interpolation framework termed WaveletVFI to address above problems. It first estimates intermediate optical flow with a lightweight motion perception network, and then a wavelet synthesis network uses flow aligned context features to predict multi-scale wavelet coefficients with sparse convolution for efficient target frame reconstruction, where the sparse valid masks that control computation in each scale are determined by a crucial threshold ratio. Instead of setting a fixed value like previous methods, we find that embedding a classifier in the motion perception network to learn a dynamic threshold for each sample can achieve more computation reduction with almost no loss of accuracy. On the common high resolution and animation frame interpolation benchmarks, proposed WaveletVFI can reduce computation up to 40\% while maintaining similar accuracy, making it perform more efficiently against other state-of-the-arts. Code is available at \url{https://github.com/ltkong218/WaveletVFI}.
\end{abstract}

\begin{IEEEkeywords}
Video frame interpolation, wavelet transform, dynamic neural networks, adaptive inference, high efficiency.
\end{IEEEkeywords}

\section{Introduction}
\IEEEPARstart{V}{ideo} frame interpolation (VFI) is an important low-level computer vision task aiming to generate non-exist intermediate frames between actual successive inputs, which can largely increase the video temporal resolution. It plays an important role in broad application prospects, such as slow motion generation~\cite{8579036}, video editing~\cite{10.1145/1186562.1015766}, animation production~\cite{Siyao_2021_CVPR} and frame rate up-conversion~\cite{7058345,6323027}.

The successful flow-based frame interpolation algorithms~\cite{Niklaus_2018_CVPR,8579036,8954114,Niklaus_2020_CVPR} can mostly be abstracted as two-stage encoder-decoder architectures, that first model optical flow between target frame and input frames, and then generate the target frame by a synthesis network. To improve the first stage, current state-of-the-arts try to adopt higher order motion model~\cite{qvi_nips19,Zhang2019video,chiall}, additional refinement unit~\cite{qvi_nips19,Sim_2021_ICCV} or directly estimate intermediate flow by a learnable network~\cite{xue2019video,Zhang_2020,huang2021rife}. As for the second stage, more powerful synthesis networks are employed to improve the frame generation ability~\cite{Niklaus_2018_CVPR,Niklaus_2020_CVPR,BMBC,park2021asymmetric}. Although significant progresses have been made by above flow-based approaches, their static deep architectures can lead to large computation redundancy on the typical piecewise flat regions in high resolution and animation videos, restricting their application scenarios to a great extent.

\begin{figure*}[t]
	\centering
	\includegraphics[width=0.98\linewidth]{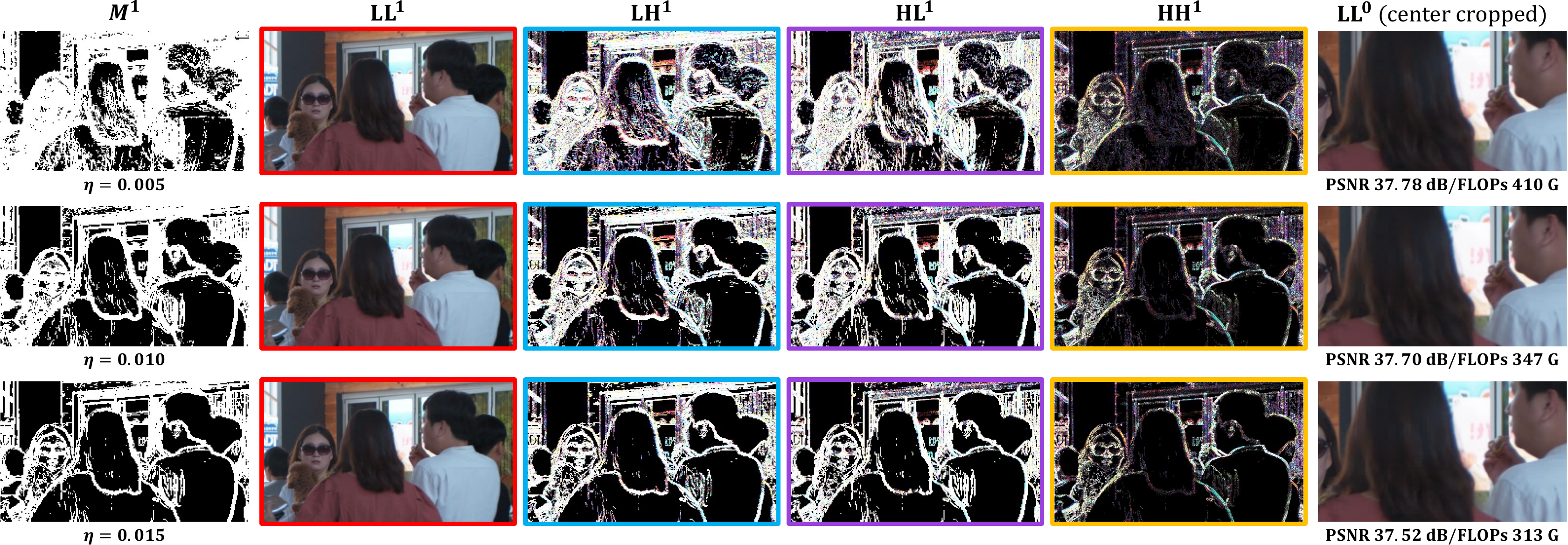}
	\caption{Target frame generation in the highest resolution decoder $\mathcal{D}_{\mathrm{WS}}^{1}$ of WS-Net. The valid mask $M^{1}$ obtained from lower level decoder $\mathcal{D}_{\mathrm{WS}}^{2}$ determines the spatial location to calculate three high-frequency wavelet coefficients $\mathrm{LH}^{1}, \mathrm{HL}^{1}, \mathrm{HH}^{1}$ with sparse convolution in decoder $\mathcal{D}_{\mathrm{WS}}^{1}$. Each row represents for synthesizing the target intermediate frame $\mathrm{LL}^{0}$ with different compression threshold ratio $\eta$, where smaller $\eta$ will take more computation cost and usually achieve better accuracy. $\mathrm{LL}^{0}$ is generated by IDWT operation applied on coefficient maps of $\mathrm{LL}^{1}, \mathrm{LH}^{1}, \mathrm{HL}^{1}$ and $\mathrm{HH}^{1}$.}
	\label{fig:1}
\end{figure*}

In this paper, inspired by the sparse representation in wavelet decomposition, we propose a novel two-stage flow-based frame interpolation algorithm called \textbf{WaveletVFI} for higher computation efficiency. Different from previous methods that directly synthesize the target frame in RGB color space~\cite{8954114,Niklaus_2020_CVPR,park2021asymmetric,Sim_2021_ICCV}, we employ discrete wavelet transform (DWT) to decompose the target frame into multi-scale frequency domain and propose a wavelet synthesis network (WS-Net) to predict the decomposed wavelet coefficients which are inherently sparse in high resolution or cartoon images. For wavelet coefficients, the low-frequency component represents the overall scene structure and the sparse high-frequency component describes some edge information. During the progressive inverse discrete wavelet transform (IDWT) based image reconstruction procedure, only the sparse high-frequency components need to be estimated in this scale. Therefore, as shown in Fig.~\ref{fig:1}, we can employ efficient sparse convolution decoder in WS-Net to predict multi-scale high-frequency wavelet coefficients only in certain areas, while still enabling high-quality intermediate frame synthesizing.

In order to build the valid spatial mask for sparse convolution, a threshold ratio has to be determined, like the quantization step in image compression~\cite{MARCELLIN200273,10.1007/978-3-030-58565-5_19}. Admittedly, the threshold hyper-parameter is an important factor that affects the computation cost and the VFI accuracy. As depicted in Fig.~\ref{fig:1}, for the same WS-Net, a lower threshold ratio $\eta$ will keep more high-frequency coefficients to be estimated, resulting in larger computation and usually higher accuracy. In contrast, more high-frequency coefficients are ignored and set to zero, often yielding lower performance while the required computation are also smaller. Thus, how to set a reasonable threshold is worth studying, that has been widely discussed in traditional image compression and denoising tasks~\cite{859238,862633}. However, in VFI task, the scene content to be generated comes from dynamic inputs, making the compression threshold for synthesizing the intermediate frame highly dependent on both texture distribution and motion situation of each input frame pair, which are difficult to be modeled explicitly. For example, given input frames with clear texture and from certain inter-frame motion, more high-frequency coefficients should be kept for better accuracy. On the other hand, when feeding input frames with blurry texture or from uncertain motion, more high-frequency coefficients can be ignored to save computation but without perceptible accuracy loss.

To deal with this problem, we propose a novel dynamic threshold ratio selection approach which can adjust the computation cost compression degree of each input sample adaptively for more efficient inference. Specifically, we introduce a threshold classifier which is embedded in the bottom part of the first stage motion perception network (MP-Net) and can learn the spatial-temporal information existed in input frames. In practice, we set several different threshold ratios as candidates and use output probability distribution over candidates of the threshold classifier as selection guidance. The MP-Net together with the threshold classifier are carefully designed with lightweight model size and computation complexity, and are jointly optimized with WS-Net for VFI task in an end-to-end manner. By exploiting the proposed dynamic compression threshold selection approach, we can better excavate computation redundancy when synthesizing the compressible target frame.

In summary, to our best knowledge, we are the first to explore the spatial redundancy problem in frame interpolation, and further build a deep VFI architecture in wavelet domain for efficient inference. Moreover, we propose a novel dynamic threshold selection mechanism to better allocate computation for each input sample. Experiments on the traditional Vimeo90K~\cite{xue2019video}, the animation ATD12K~\cite{Siyao_2021_CVPR} and the high resolution Xiph-2K~\cite{Montgomery1994Xiph} and Xiph-4K~\cite{Montgomery1994Xiph} frame interpolation benchmarks demonstrate the effectiveness of proposed approaches, which can adaptively reduce the overall resource consumption while maintaining advanced VFI accuracy.

\section{Related Work}
\subsection{Video Frame Interpolation}
Research in deep learning based frame interpolation can be roughly categorized into flow-based and kernel-based approaches. Kernel-based methods adopt adaptive convolution~\cite{8099727}, where they unify motion estimation and frame generation into a single convolution step with spatial varying convolution kernels. Following works mainly enhance the freedom of convolution operation~\cite{8237299,Lee_2020_CVPR,Cheng_Chen_2020}, combining optical flow offsets for better spatial alignment~\cite{Gui_2020_CVPR}, or introducing channel attention mechanism~\cite{choi2020cain}. Kernel-based approaches can naturally generate complex contextual details, however, their prediction tend to be blurry when scene motion is large.

Recent state-of-the-arts mostly adopt flow-based methods~\cite{8579036,8954114,qvi_nips19,Niklaus_2020_CVPR,Siyao_2021_CVPR,park2021asymmetric,Sim_2021_ICCV,Kong_2022_CVPR,Liu_2022_ICIP,9944840}, since optical flow can provide an explicit correspondence for frame registration, especially in large motion scenes. Due to there is an independent motion modeling step in flow-based approaches, they usually contain a second target frame synthesizing stage. Existing improvement for the first stage try to adopt more advanced motion model~\cite{qvi_nips19,Zhang2019video,chiall,Liu_2022_ICIP}, additional refinement unit~\cite{qvi_nips19,Sim_2021_ICCV} or directly estimate intermediate flow by an encoder-decoder network~\cite{xue2019video,Zhang_2020,huang2021rife}. As for the second stage, more powerful synthesis networks are employed to improve the frame generation ability~\cite{Niklaus_2018_CVPR,Niklaus_2020_CVPR,BMBC,park2021asymmetric}.

In order to achieve more efficiency, CDFI~\cite{ding2021cdfi} first leverages model pruning through sparsity-inducing optimization, and then add additional synthesis module to improve previous compressed network, which significantly reduces model size against the baseline AdaCoF~\cite{Lee_2020_CVPR}. However, their complex architecture leads to large time delay and the results are inferior to current SOTA methods. Like our approach, CAIN-SD~\cite{Choi_2021_ICCV} also adopts a motion-aware dynamic architecture for efficiency, where they dynamically adjust the network depth and input resolution for each input image patch. However, their stitched target frame contains artifacts at patch edges, and their base model CAIN~\cite{choi2020cain} can not deal with large motion as well as flow-based VFI methods. Recently, IFRNet~\cite{Kong_2022_CVPR} achieves state-of-the-art speed accuracy trade-off by jointly refining intermediate optical flow together with a powerful intermediate feature within a single encoder-decoder architecture. However, the fully convolutional structure treats each pixel for synthesizing equally, that can result in large computation redundancy when generating lots of smooth regions.

\subsection{Wavelets in Computer Vision}
Wavelet decomposition and reconstruction is widely used in signal processing, image processing and computer vision. The discrete wavelet transform (DWT) can make the signal energy distribution more concentrated on principle frequency components and hence compress redundant information. The JPEG2000~\cite{1221761,10.5555/2588198} standard employs DWT algorithm in the image compression stage, and a truncated threshold plays a key role for quantization~\cite{859238,862633}. The frequency filtering characteristic of wavelet transform is also applied on traditional image denoising task~\cite{382009,862633}.

Recently, wavelet transform has been combined with diverse deep learning based computer vision tasks. WDNet~\cite{10.1007/978-3-030-58601-0_6} proposes to remove image moiré artifacts in the wavelet domain, which is difficult to distinguish from true texture in the RGB color space. Some super-resolution methods~\cite{8237449,10.1145/3394171.3413664,9008402} learn to estimate multiple high-frequency wavelet coefficients from a low-resolution input image to generate high-resolution image by inverse discrete wavelet transform (IDWT). WaveletStereo~\cite{Yang_2020_CVPR} learns stereo matching by predicting the wavelet coefficients of the disparity, that can better deal with global context with textureless surfaces. Closer to our work, WaveletMonodepth~\cite{Ramamonjisoa_2021_CVPR} predicts multi-scale sparse wavelet coefficients for efficient monocular depth estimation. However, their compression threshold value can not dynamically adjust on every sample for better accuracy efficiency trade-off. Moreover, the threshold selection in dynamic VFI task is more complex than the static monocular depth task, since motion uncertainty will influence the compression characteristic curve. To our best knowledge, we are the first to apply wavelet transform to frame interpolation, and further in a dynamic manner.

\begin{figure*}[t]
	\centering
	\includegraphics[width=0.98\linewidth]{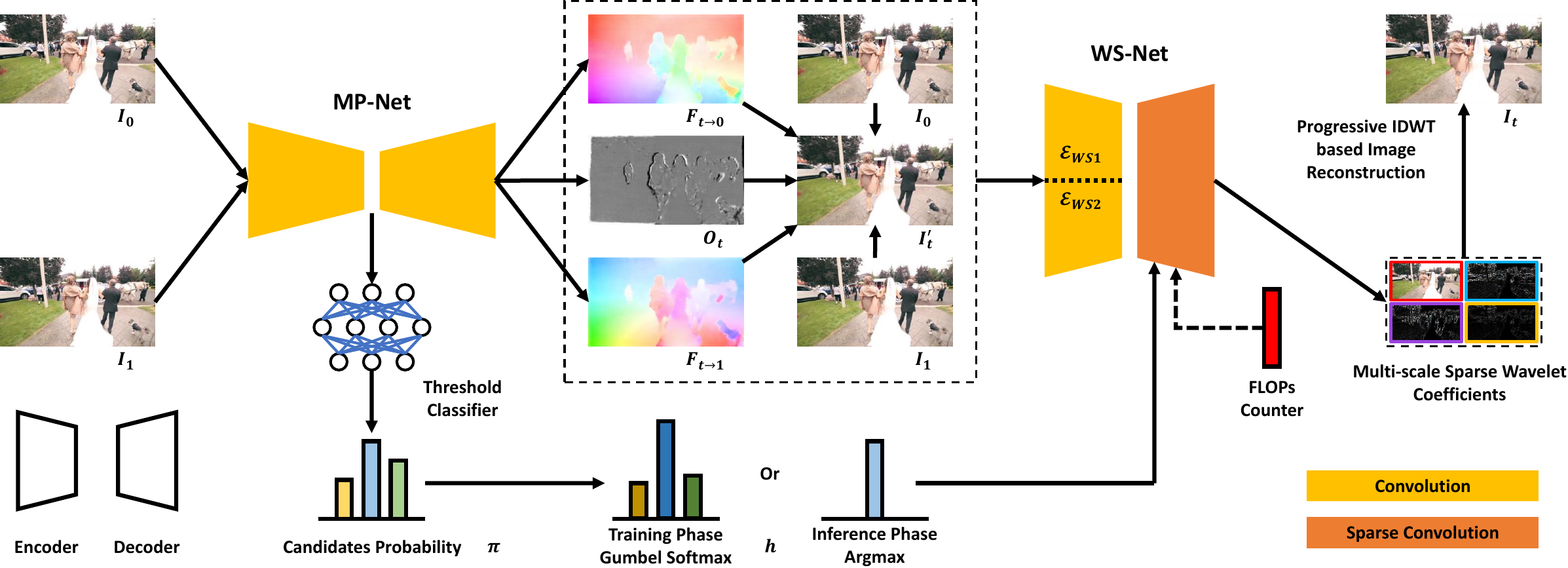}
	\caption{Overall framework of our WaveletVFI that can interpolate frames dynamically in wavelet domain. It contains a motion perception network (MP-Net) and a wavelet synthesis network (WS-Net), where the first model estimates intermediate optical flow and occlusion merge mask and the second network encodes diverse spatial aligned inputs and predicts multi-scale sparse wavelet coefficient maps for progressive IDWT based target frame reconstruction. The compression threshold classifier is a lightweight neural network which is embedded into the MP-Net to perceive spatial-temporal input and select adaptive compression threshold ratio for adjusting computation cost. By leveraging the Gumbel softmax trick~\cite{NIPS2014_309fee4e,Jang2017CategoricalRW}, proposed WaveletVFI can be trained end-to-end.}
	\label{fig:2}
\end{figure*}

\subsection{Dynamic Neural Networks}
Dynamic neural networks, as opposed to traditional static models, can adapt their structures or parameters according to the input during inference, and therefore enjoy favorable properties that are absent in static ones. One of the most notable advantages of dynamic models is that they are able to allocate computations on demand in test time. Common practices mostly include dynamic depth, dynamic width and spatial-wise dynamic networks. The dynamic depth approaches contain two major types of early exiting~\cite{Bolukbasi_2017,huang2018multi} and layer skipping~\cite{wang2018skipnet,Veit2018}. The dynamic width networks usually skip neurons in fully-connected (FC) layers~\cite{Bengio2013EstimatingOP,Bengio2015ConditionalCI} or skip channels in convolutional neural networks (CNNs)~\cite{NIPS2017_a51fb975,Li_2021_CVPR}. The spatial-wise dynamic networks often leverage dynamic sparse convolution to reduce the unnecessary computation on less informative locations, where our WaveletVFI falls into this category. To determine the spatial location for sparse convolution, diverse sampling strategies are invented. A typical approach is to use an extra network branch to generate a spatial valid mask where many methods~\cite{8954289,8658916,Verelst_2020_CVPR} belong to this paradigm. Another kind algorithms make use of the sparse characteristics of the input~\cite{8579006}. Different from these methods, our approach leverages the intrinsic sparsity of wavelet representation and explore the optimal sparsity degree by learning from the spatial-temporal motion information in an end-to-end manner, that is especially suitable for frame interpolation task.

\section{Method}
In this section, we first introduce the overall framework of the proposed method. Then, we describe the bidirectional intermediate flow estimation and dynamic compression threshold selection approach in motion perception network. Further, complementary context encoder, sparse convolution decoder and progressive IDWT based target frame reconstruction algorithm in wavelet synthesis network are demonstrated. Finally, we present the optimization procedure and loss functions.

\subsection{Framework Overview}
Inspired by the sparse representation characteristic of wavelet decomposition and threshold selection methods on image compression~\cite{1221761,859238,862633}, we propose an instance-aware dynamic compression threshold selection approach in wavelet domain for efficient video frame interpolation. As shown in Fig.~\ref{fig:2}, our proposed method follows the successful two-stage flow-based VFI pipeline. In the first stage, the MP-Net jointly estimates bidirectional intermediate flow $F_{t\rightarrow0}, F_{t\rightarrow1}$ and an occlusion fusion mask $O_t$ from coarse to fine. Then, the intermediate flow, occlusion mask, input images $I_0, I_1$ and merged intermediate frame $I_{t}^{'}$ are fed into the encoder part of WS-Net in the second stage. Meanwhile, the threshold classifier in MP-Net learns the spatial-temporal input information and provides candidates probability for compression threshold selection, which controls the computation cost and synthesis accuracy of the sparse convolution decoder part of WS-Net.  In training, candidates probability is applied with Gumbel softmax trick~\cite{NIPS2014_309fee4e,Jang2017CategoricalRW} for gradient back propagation, while in inference, only threshold ratio with maximum probability is selected for synthesizing.

\subsection{Motion Perception Network}
\begin{figure*}[t]
	\centering
	\includegraphics[width=0.98\linewidth]{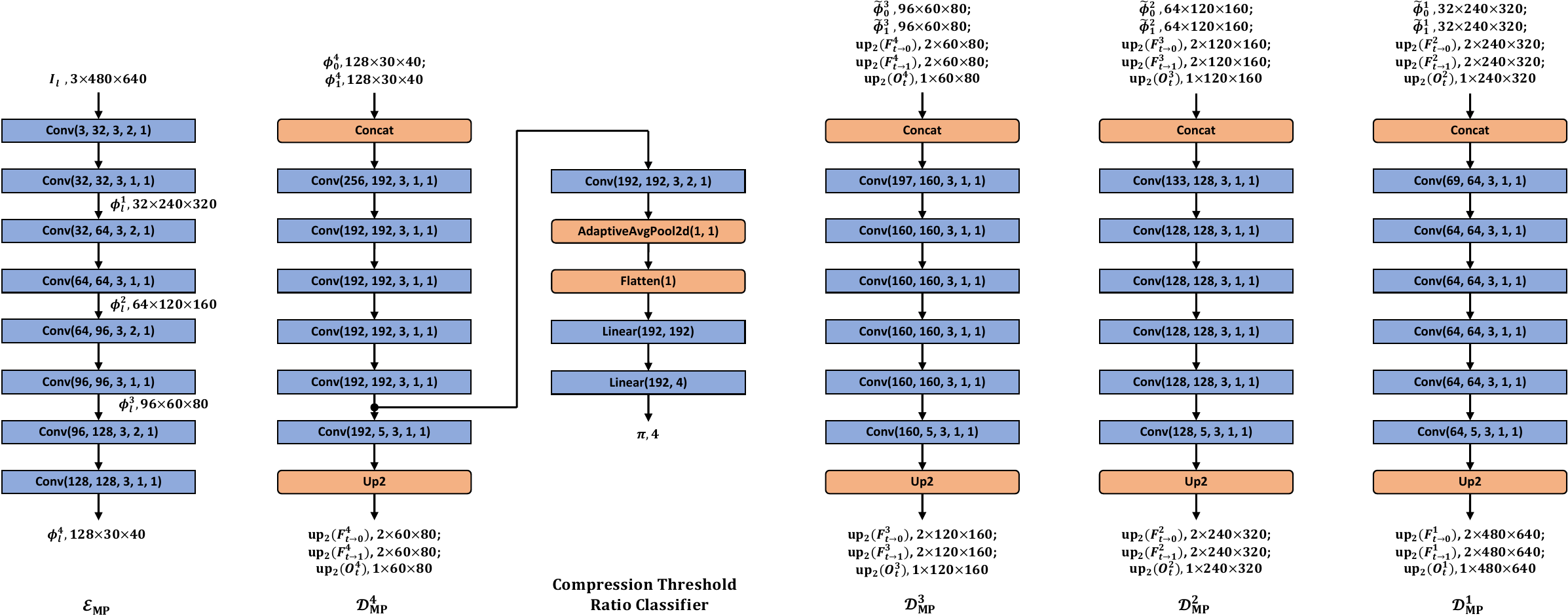}
	\caption{Structure details of the pyramid encoder $\mathcal{E}_{\mathrm{MP}}$ and coarse-to-fine decoders $\mathcal{D}_{\mathrm{MP}}^{4}, \mathcal{D}_{\mathrm{MP}}^{3}, \mathcal{D}_{\mathrm{MP}}^{2}, \mathcal{D}_{\mathrm{MP}}^{1}$ in proposed motion perception network $\mathcal{N}_{\mathrm{MP}}$. The compression threshold ratio classifier is a branch of $\mathcal{D}_{\mathrm{MP}}^{4}$. Arguments of `Conv' from left to right are input channels, output channels, kernel size, stride and padding, respectively. Dimensions of input and output tensors from left to right stand for feature channels, height and width, separately. A Leaky ReLU activation with negative slope set to 0.1 follows each learnable layer except for the last one. We take input frames with resolution 640 $\times$ 480 as example.}
	\label{fig:3}
\end{figure*}

\subsubsection{Joint Intermediate Flow and Occlusion Estimation}
The MP-Net takes two input frames $I_0, I_1$ and jointly estimates bidirectional intermediate optical flow $F_{t\rightarrow0}, F_{t\rightarrow1}$ with occlusion fusion mask $O_t$ in a coarse-to-fine manner. Specifically, the pyramid encoder of MP-Net extracts 4 levels of pyramid features, \textit{i.e.}, $\phi_0^1, \phi_0^2, \phi_0^3, \phi_0^4$ and $\phi_1^1, \phi_1^2, \phi_1^3, \phi_1^4$ from $I_0$ and $I_1$ respectively, where the spatial resolution of level $l+1$ is $1/2$ of level $l$. The bottom level decoder $\mathcal{D}_{\mathrm{MP}}^4$ directly takes the concatenation of $\phi_0^4, \phi_1^4$ as input, and estimates a coarse intermediate flow $F_{t\rightarrow0}^4, F_{t\rightarrow1}^4$ and fusion mask $O_t^4$. Following the success of pyramid methods~\cite{8579029,9191101,9413531} in large displacement optical flow estimation, we adopt the 2 $\times$ upsampled intermediate flow $\mathtt{up}_2(F_{t\rightarrow0}^4), \mathtt{up}_2(F_{t\rightarrow1}^4)$ to backward warp pyramid features $\phi_0^3, \phi_1^3$ and obtained the warped features $\tilde{\phi}_{0}^{3}, \tilde{\phi}_{1}^{3}$ respectively. Then, the concatenated features of $\mathtt{up}_2(F_{t\rightarrow0}^4), \mathtt{up}_2(F_{t\rightarrow1}^4), \mathtt{up}_2(O_t^4)$ and $\tilde{\phi}_{0}^{3}, \tilde{\phi}_{1}^{3}$ are fed to decoder $\mathcal{D}_{\mathrm{MP}}^3$ for estimating finer intermediate flow $F_{t\rightarrow0}^3, F_{t\rightarrow1}^3$ and occlusion mask $O_t^3$. This procedure is performed recursively until reaching the original input resolution and yielding $F_{t\rightarrow0}, F_{t\rightarrow1}, O_t$. In experiments, we find that integrate occlusion mask $O_t$ with intermediate flow $F_{t\rightarrow0}, F_{t\rightarrow1}$ for joint refinement can provide more useful information for the following synthesis network with negligible additional cost. Concretely, we can build a merged intermediate frame $I_{t}^{'}$ to better guide the following WS-Net by
\begin{align}
	I_{t}^{'} = O_t \odot \tilde{I}_{0} & + (1 - O_t) \odot \tilde{I}_{1}, \\
	\tilde{I}_{0} = \mathtt{warp}(I_{0}, F_{t\rightarrow0}) & , \; \tilde{I}_{1} = \mathtt{warp}(I_{1}, F_{t\rightarrow1}),
\end{align}
where $\mathtt{warp}$ means backward warping, $\odot$ stands for element-wise multiplication, and $-$ is element-wise subtraction. Fig.~\ref{fig:3} shows structure details of the motion perception network $\mathcal{N}_{\mathrm{MP}}$.

\begin{figure*}[t]
	\centering
	\includegraphics[width=0.8\linewidth]{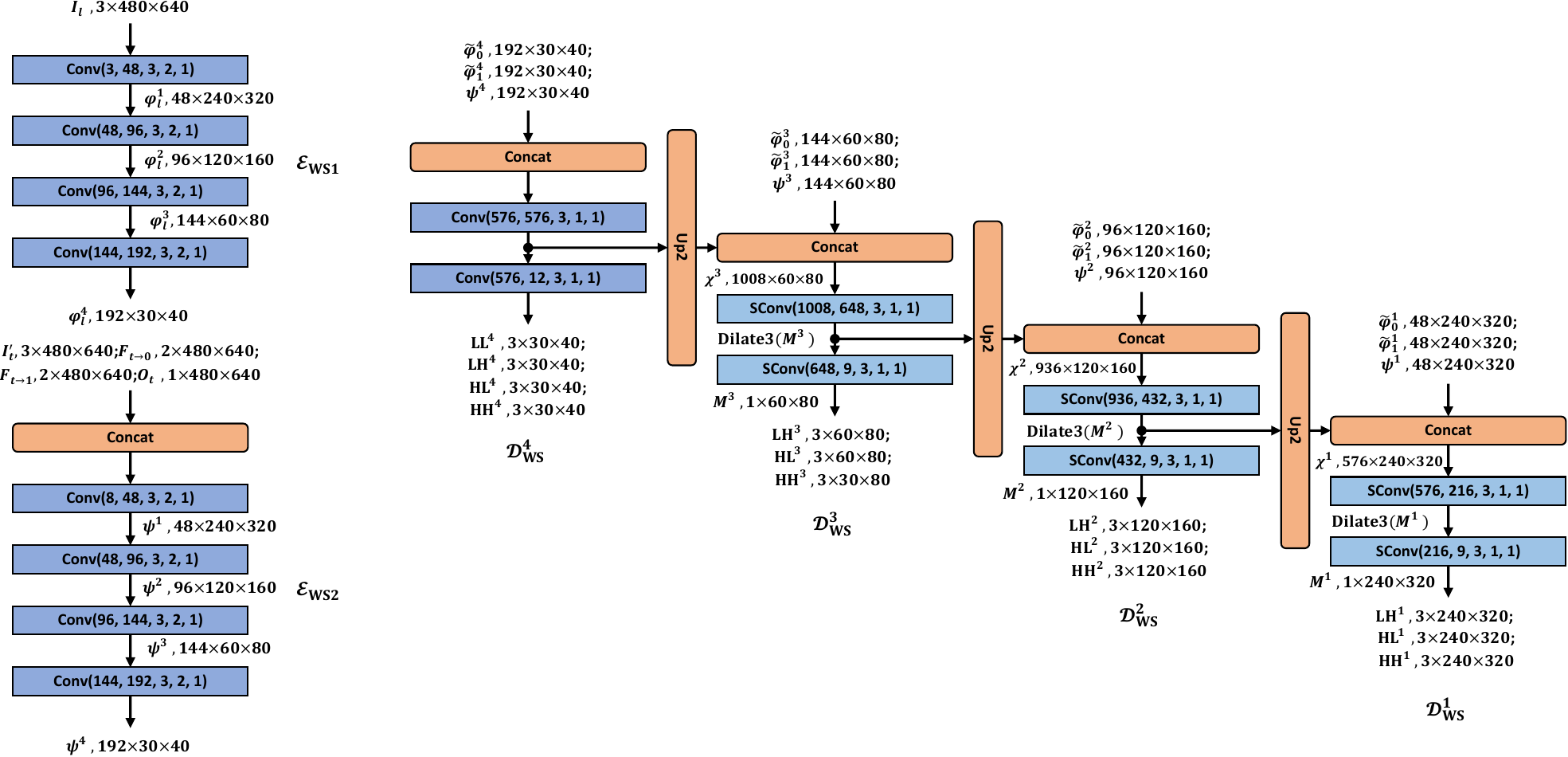}
	\caption{Structure details of the complementary context encoders $\mathcal{E}_{\mathrm{WS1}}, \mathcal{E}_{\mathrm{WS2}}$ and coarse-to-fine decoders $\mathcal{D}_{\mathrm{WS}}^{4}, \mathcal{D}_{\mathrm{WS}}^{3}, \mathcal{D}_{\mathrm{WS}}^{2}, \mathcal{D}_{\mathrm{MP}}^{1}$ in proposed wavelet synthesis network $\mathcal{N}_{\mathrm{WS}}$. `SConv' means sparse convolution. Arguments of `Conv' and `SConv' from left to right are input channels, output channels, kernel size, stride and padding, respectively. Dimensions of input and output tensors from left to right stand for feature channels, height and width, separately. A Leaky ReLU activation with negative slope set to 0.1 follows each learnable layer except for the last one. We take input resolution 640 $\times$ 480 as example.}
	\label{fig:4}
\end{figure*}

\subsubsection{Compression Threshold Classifier}
The role of our proposed compression threshold ratio classifier, abbreviated as threshold classifier, is to decide the threshold ratio hyper-parameter $\eta$ for the following WS-Net, which controls the trade-off between computation complexity and target frame synthesis accuracy. However, different from the image compression~\cite{10.5555/2588198,859238} and monocular depth estimation~\cite{Ramamonjisoa_2021_CVPR} tasks in wavelet domain, where the compression threshold is mainly affected by static scene content, in frame interpolation, the compression degree of target frame is influenced by both scene structure and motion situation, which are more complex to model. For example, target frames synthesized from input samples with blur, exposure and other noisy texture in the challenging motion scenes usually contain more unreliable high-frequency texture, which are more compressible and can even achieve better results by the denoising characteristics of wavelet transform~\cite{803428,862633}. On the other hand, target frame with rich texture and from certain motion should keep more high-frequency wavelet coefficients for better quantitative and qualitative results. To deal with above intractable problem, we introduce the threshold classifier in MP-Net to find an appropriate instance-aware threshold ratio by inferring a probability distribution over candidate threshold ratios. In practice, as shown in Fig.~\ref{fig:3}, the threshold classifier is a lightweight network with one convolution and two fully-connected layers separated by Leaky ReLU activation, that is embedded to the second last convolution layer of decoder $\mathcal{D}_{\mathrm{MP}}^{4}$ in MP-Net. Taking $m$ threshold ratios of $\eta_1, \eta_2, ..., \eta_m$ as candidates, the threshold classifier predicts a categorical distribution $\pi = [\pi_1, \pi_2, ..., \pi_m]$ over them. Since there exists an non-differentiable problem in the process from the soft probability outputs $\pi$ to the hard one-hot selection $h\in\{0, 1\}^m$, we leverage the Gumbel softmax trick~\cite{NIPS2014_309fee4e,Jang2017CategoricalRW} to make the discrete decision differentiable during the gradient back propagation, which means the discrete candidate threshold selections can be drawn by using
\begin{equation}
	h = \mathtt{one\_hot}[\mathop{\arg\max}_{k}(\mathrm{log} (\pi_k) + g_k)], 
\end{equation}
where $g_k \sim \mathrm{Gumbel}(0, 1)$ is an i.i.d Gumbel noise sample, which will not influence the highest entry of the original categorical probability distribution. During training, the derivative of above one-hot operation can be approximated by Gumbel softmax function that is both continuous and differentiable
\begin{equation}
	h_k = \frac{\mathrm{exp}[(\mathrm{log}(\pi_k) + g_k) / \tau]}{\sum_{j=1}^{m} \mathrm{exp}[(\mathrm{log}(\pi_j) + g_j) / \tau]}, 
	\label{eq:4}
\end{equation}
where $\tau$ is a temperature parameter. When $\tau \rightarrow \infty$, samples from Gumbel softmax distribution become uniform. In contrast, when $\tau \rightarrow 0$, samples from Gumbel softmax distribution become one-hot. In our experiments, we start at a high temperature of $\tau=1.0$ and anneal it to $0.4$ finally.

\begin{figure}[t]
	\centering
	\includegraphics[width=0.96\linewidth]{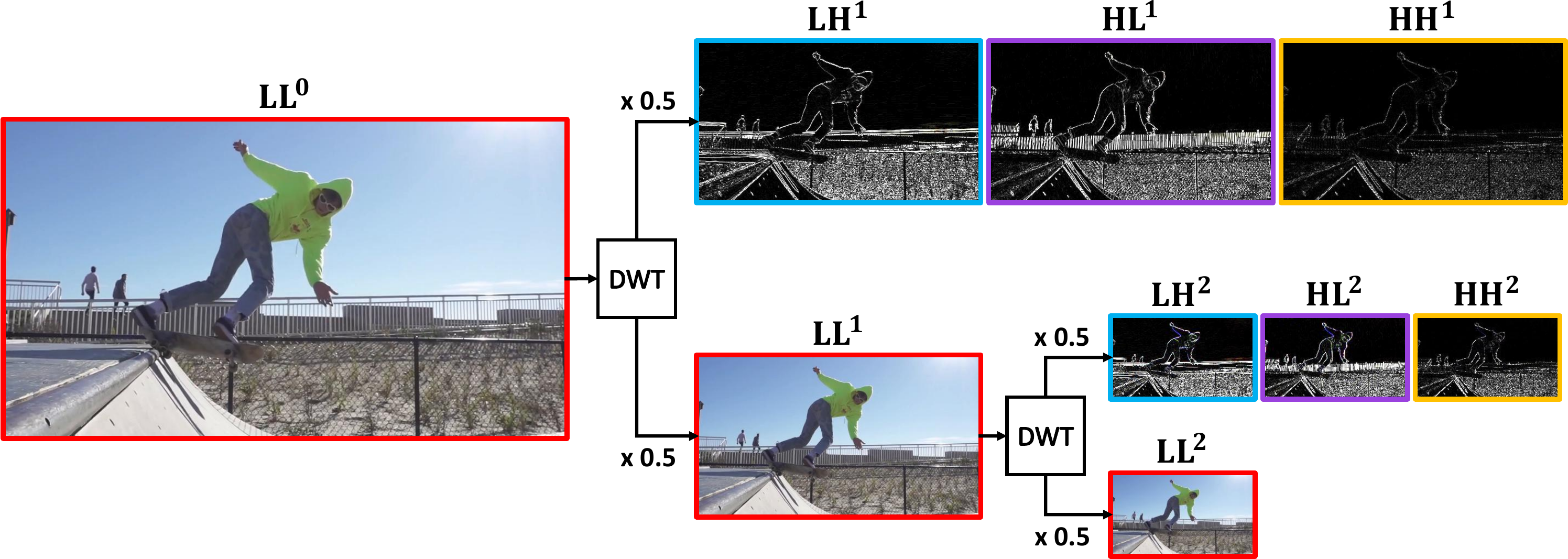}
	\caption{Progressive DWT with Haar kernels for image decomposition.}
	\label{fig:5}
\end{figure}

\begin{figure}[t]
	\centering
	\includegraphics[width=0.96\linewidth]{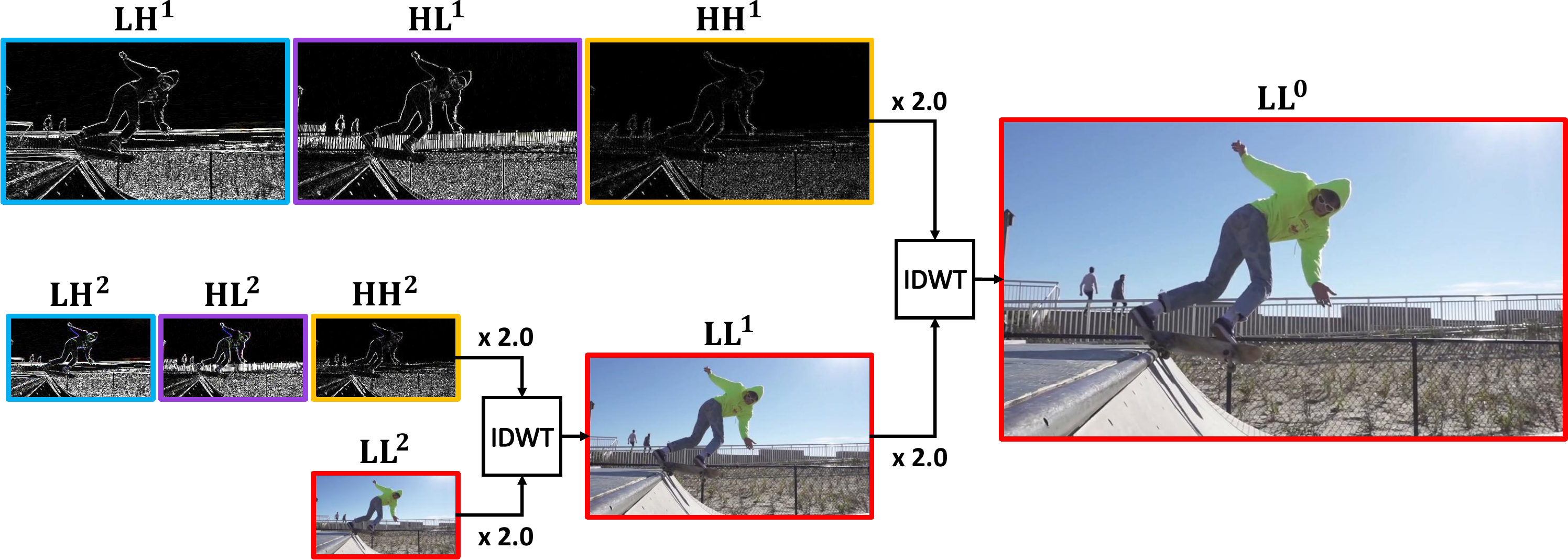}
	\caption{Progressive IDWT with Haar kernels for image reconstruction.}
	\label{fig:6}
\end{figure}

\subsection{Wavelet Synthesis Network}
\subsubsection{Complementary Context Encoder}
Like MP-Net, the WS-Net is also a U-shape encoder-decoder network, however, it has more feature channels but with less cascaded convolutions in each scale. The reason is that flow warped context features have almost been aligned to the target position, while more feature channels are needed for encoding diverse image texture. Different from previous methods~\cite{Niklaus_2018_CVPR,Niklaus_2020_CVPR,huang2021rife} that adopt only a single encoder, we employ two different encoders $\mathcal{E}_{\mathrm{WS1}}, \mathcal{E}_{\mathrm{WS2}}$ to extract complementary context features. Specifically, $\mathcal{E}_{\mathrm{WS1}}$ extracts 4 levels of pyramid features of $\varphi_0^1, \varphi_0^2, \varphi_0^3, \varphi_0^4$ and $\varphi_1^1, \varphi_1^2, \varphi_1^3, \varphi_1^4$ from $I_0$ and $I_1$ separately. Then, they are warped by progressively down sampled intermediate flow fields $F_{t\rightarrow0}, F_{t\rightarrow1}$ to obtain target frame aligned context features of $\tilde{\varphi}_{0}^{l}, \tilde{\varphi}_{1}^{l}, l\in\{1, 2, 3, 4\}$ respectively. On the other hand, $\mathcal{E}_{\mathrm{WS2}}$ takes concatenation of $F_{t\rightarrow0}, F_{t\rightarrow1}, O_t$ and $I_{t}^{'}$ as input, and also extracts 4 levels of pyramid features of $\psi^l, l\in\{1, 2, 3, 4\}$, that contain additional scene motion and occlusion information. Finally, we take the concatenated features of $\tilde{\varphi}_{0}^{l}, \tilde{\varphi}_{1}^{l}$ and $\psi^l$ as the complementary context feature in each level $l$. Details of $\mathcal{E}_{\mathrm{WS1}}, \mathcal{E}_{\mathrm{WS2}}$ are depicted in Fig.~\ref{fig:4}.

\subsubsection{Haar Wavelet Transform}
Before introducing sparse convolution decoder, we first explain the Haar wavelets used in our WaveletVFI, which has the simplest basis functions in discrete wavelet transform (DWT). Haar wavelet transform has four kernels, \textit{i.e.}, $\{\mathrm{LL}^{\top}, \mathrm{LH}^{\top}, \mathrm{HL}^{\top}, \mathrm{HH}^{\top}\}$, where the low ($\mathrm{L}$) and high ($\mathrm{H}$) pass filters are
\begin{equation}
	\mathrm{L}^{\top} = \frac{1}{\sqrt{2}}[1 \; \; 1], \quad \mathrm{H}^{\top} = \frac{1}{\sqrt{2}}[-1 \; \; 1].
\end{equation}
DWT with Haar wavelets can decompose a 2D image into four coefficient maps, including a low-frequency component $\mathrm{LL}$ and three high-frequency components $\mathrm{LH, HL, HH}$ at half the resolution of input image, where $\mathrm{LL}$ captures smooth texture while $\mathrm{LH, HL, HH}$ extract vertical, horizontal and diagonal `jump' information. Since DWT is an invertible operation, we can adopt its inverse, \textit{i.e.} IDWT, to convert four coefficient maps back to the 2D image at double the resolution of coefficient maps. To extract multi-scale and multi-frequency wavelet representation from ground truth intermediate frame $\hat{I}_t$, we can apply DWT operation recursively on the low-frequency coefficient map $\hat{\mathrm{LL}}$, starting from the input image $\hat{I}_t$, as shown in Fig.~\ref{fig:5}. Correspondingly, to reconstruct the predicted target frame $I_{t}$, decoders in WS-Net $\mathcal{D}_{\mathrm{WS}}^{l}$, except the bottom one $\mathcal{D}_{\mathrm{WS}}^{4}$, only need to estimate three high-frequency wavelet coefficients in this scale, and apply IDWT operation on them recursively until reaching the original input resolution, that is shown in Fig.~\ref{fig:6}. Formally, these two mutually inverse transforms can be written as
\begin{align}
	&\hat{\mathrm{LL}}^{l}, \hat{\mathrm{LH}}^{l}, \hat{\mathrm{HL}}^{l}, \hat{\mathrm{HH}}^{l} \leftarrow \mathrm{DWT}(\hat{\mathrm{LL}}^{l-1}), \\
	&\mathrm{LL}^{l-1} \leftarrow \mathrm{IDWT}(\mathrm{LL}^{l}, \mathrm{LH}^{l}, \mathrm{HL}^{l}, \mathrm{HH}^{l}),
\end{align}
where superscript $l, l\in \{1, 2, 3, 4\}$ denotes the current pyramid level, $\hat{}$ means the ground truth wavelet coefficients, $\hat{\mathrm{LL}}^{0}$ and $\mathrm{LL}^{0}$ equals to ground truth frame $\hat{I}_{t}$ and predicted target frame $I_t$ respectively.

\subsubsection{Sparse Convolution Decoder}
For the piecewise flat regions in high resolution and cartoon images, most of their high-frequency wavelet coefficients have small values that are close to zero, while only some noticeable values are around image edges. Therefore, for full-resolution target frame reconstruction, only certain pixel locations need to estimate non-zero wavelet coefficients at each scale. Denoting these certain locations as sparse valid mask $M^{l} \in \{0, 1\}^{H^l \times W^l}$ in level $l$, where $1$ means valid, we can exploit sparse convolution to build decoder $\mathcal{D}_{\mathrm{WS}}^{l}$ for efficient calculation as
\begin{equation}
	\mathrm{LH}^{l}, \mathrm{HL}^{l}, \mathrm{HH}^{l} = \mathcal{D}_{\mathrm{WS}}^{l}(\chi^{l}, M^{l}), \; l \in \{1, 2, 3\}, 
\end{equation}
where $\chi^{l}$ stands for the concatenated encoding and decoding pyramid features in level $l$ of the U-shape WS-Net. $M^{l}$ denotes the sparse valid mask of the last sparse convolution in decoder $\mathcal{D}_{\mathrm{WS}}^{l}$. To get meaningful value during sparse inference, we use the 3$\times$3 morphological dilate operation $\mathtt{dilate}_{3}$ to obtain the sparse valid mask of the first sparse convolution in each decoder $\mathcal{D}_{\mathrm{WS}}^{l}$. Given predicted multi-scale sparse wavelet coefficients, we can get the predicted intermediate frame $I_t$, \textit{i.e.}, $\mathrm{LL}^{0}$ by exploiting inverse discrete wavelet transform (IDWT) progressively. It is worth noting that elements in the initial valid mask $M^{4}$ are all set to $1$, and $\mathcal{D}_{\mathrm{WS}}^{4}$ predicts an additional low-frequency coefficient $\mathrm{LL}^{4}$.

\subsubsection{Sparse Valid Mask Calculation}
Finally, we demonstrate how to calculate sparse valid mask $M^{l} $ in level $l, l\in \{1, 2, 3\}$, where the compression threshold ratio $\eta$ plays a key role as previously discussed. Inspired by the spatial correlation of different wavelet coefficient maps among multiple scales, which is first raised in the zerotree wavelets encoding algorithm~\cite{136601}, we assume that $M^{l}$ can be determined with high-frequency coefficient maps estimated at the previous scale by
\begin{align}
	M^{l} = \mathtt{up}_2(\mathrm{max}(|\mathrm{LH}^{l+1}|, |\mathrm{HL}^{l+1}|, | & \mathrm{HH}^{l+1}|) > \nonumber \\
	\eta \cdot (\mathrm{max}(\mathrm{LL}^{l}) & - \mathrm{min}(\mathrm{LL}^{l}))).
	\label{eq:9}
\end{align}
Since the target frame $I_t$ is a 3-channel RGB image, we first calculate valid mask $M_{c}^{l}$ for each color channel, then we take the union set of $M_{c}^{l}, c\in \{R, G, B\}$ as the final $M^{l}$. As shown in Eq.~\ref{eq:9}, a larger $\eta$ will make $M^{l}$ more sparse, which usually leads to less computation and lower synthesis accuracy.

\begin{algorithm}[t]
	\small
	\SetAlgoLined
	\caption{Progressive Target Frame Reconstruction}
	\KwIn{Pyramid features: [$\chi^4, \chi^3, \chi^2, \chi^1$]; Compression threshold ratio: $\eta$.}
	\KwOut{Predicted intermediate frame: $\mathrm{LL}^0$; Predicted multi-scale wavelet coefficients set: $\mathbb{W} = \{\mathrm{LL}^{l}, \mathrm{LH}^{l}, \mathrm{HL}^{l}, \mathrm{HH}^{l} | l=1, 2, 3, 4\}$.}
	\BlankLine
	\setstretch{1.1}
	$\mathrm{LL}^{4}, \mathrm{LH}^{4}, \mathrm{HL}^{4}, \mathrm{HH}^{4} = \mathcal{D}_{\mathrm{WS}}^{4}(\chi^{4})$;\\
	$\mathrm{LL}^{3} \leftarrow \mathrm{IDWT}(\mathrm{LL}^{4}, \mathrm{LH}^{4}, \mathrm{HL}^{4}, \mathrm{HH}^{4})$;\\
	\setstretch{1.3}
	\For{$l=3;\ l>0;\ l=l-1$}{
		$\eta^l = \eta \cdot (\mathrm{max}(\mathrm{LL}^{l}) - \mathrm{min}(\mathrm{LL}^{l}))$;\\
		$M^{l} = \mathtt{up}_2(\mathrm{max}(|\mathrm{LH}^{l+1}|, |\mathrm{HL}^{l+1}|, |\mathrm{HH}^{l+1}|) > \eta^l$);\\
		$\mathrm{LH}^{l}, \mathrm{HL}^{l}, \mathrm{HH}^{l} = \mathcal{D}_{\mathrm{WS}}^{l}(\chi^{l}, M^{l})$;\\
		$\mathrm{LL}^{l-1} \leftarrow \mathrm{IDWT}(\mathrm{LL}^{l}, \mathrm{LH}^{l}, \mathrm{HL}^{l}, \mathrm{HH}^{l})$;\\
	}
	\label{algo:1}
\end{algorithm}

In summary, the target frame synthesis procedure in proposed WS-Net involves sparse valid mask calculation, sparse convolution inference and progressive IDWT based image reconstruction, we arrange these approaches into an algorithm for clarity which is presented in Algorithm~\ref{algo:1}.

\subsection{Optimization}
\subsubsection{Differentiable Forward Propagation}
Based on above analysis, the forward propagation stage of WaveletVFI can be summarized as following steps: \textbf{1)} Given two input frames $I_0, I_1$, $\mathcal{N}_{\mathrm{MP}}$ predicts $F_{t\rightarrow0}, F_{t\rightarrow1}, O_t$ and a discrete candidate threshold selection $h$ by Gumbel softmax trick in Eq.~\ref{eq:4}. \textbf{2)} Given $I_0, I_1, F_{t\rightarrow0}, F_{t\rightarrow1}, O_t, I_{t}^{'}$ and a specific candidate selection $h_k, k \in \{1, 2, ..., m\}$, $\mathcal{N}_{\mathrm{WS}}$ estimates the multi-scale wavelet coefficients set $\mathbb{W}_{k}$, which is abbreviated as $\mathbb{W}_{k} = \mathcal{N}_{\mathrm{WS}}(\cdots; h_k)$. \textbf{3)} Denoting above progressive IDWT based target frame reconstruction algorithm as $\mathcal{A}_{\mathrm{IDWT}}$, the predicted intermediate frame $\mathrm{LL}_{k}^{0}$ based on the $k$-th candidate selection $h_k$ can be obtained by $\mathrm{LL}_{k}^{0} = \mathcal{A}_{\mathrm{IDWT}}(\mathbb{W}_{k})$. \textbf{4)} Given the predicted candidate selection $h$, we can get the final predicted target frame $\mathrm{LL}^{0}$ and multi-scale wavelet coefficients set $\mathbb{W}$ by summing up $h_k$ with $h$ as follows
\begin{align}
	\mathrm{LL}^{0} = & {\sum}_{k=1}^{m} h_k \cdot  \mathcal{A}_{\mathrm{IDWT}}(\mathcal{N}_{\mathrm{WS}}(\cdots; h_k)), \\[2mm]
	& \mathbb{W} = {\sum}_{k=1}^{m} h_k \cdot \mathcal{N}_{\mathrm{WS}}(\cdots; h_k).
\end{align}

\begin{figure*}[t]
	\captionsetup[subfloat]{labelformat=empty,captionskip=-1pt}
	\begin{center}
		\newcommand{\rowArg}{2.11cm}
		\newcommand{\fullHeight}{4.27cm}
		\newcommand{\fullWidth}{7.59cm}
		\newcommand{\patchSize}{1.96cm}
		\scriptsize
		\setlength\tabcolsep{0.05cm}
		\begin{tabular}[b]{c c c c c c}
			\multirow{2}{*}[\rowArg]{
				\subfloat[\scriptsize Ground Truth Frame from ATD12K~\cite{Siyao_2021_CVPR}]
				{\includegraphics[width = \fullWidth, height = \fullHeight]
					{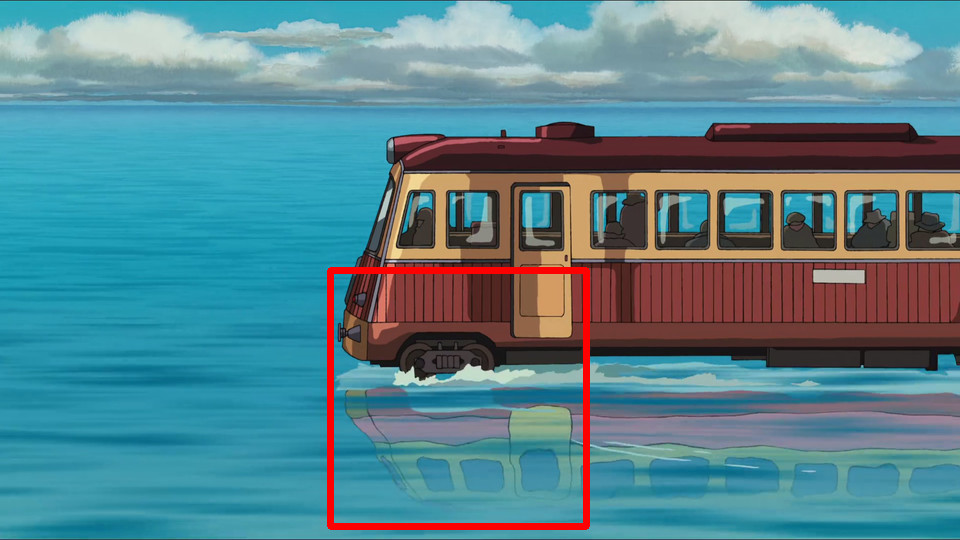}}} &
			\subfloat[\scriptsize Overlay]
			{\includegraphics[width = \patchSize, height = \patchSize]
				{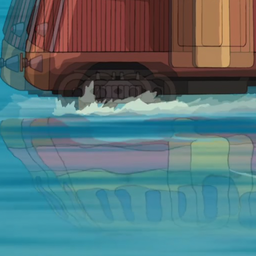}} &
			\subfloat[\scriptsize SepConv~\cite{8237299}]
			{\includegraphics[width = \patchSize, height = \patchSize]
				{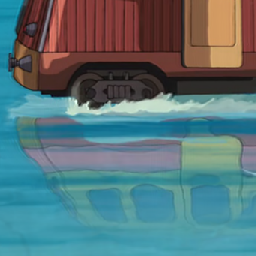}} &
			\subfloat[\scriptsize DAIN~\cite{8954114}]
			{\includegraphics[width = \patchSize, height = \patchSize]
				{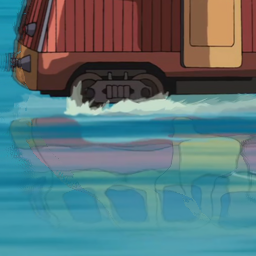}} &
			\subfloat[\scriptsize CAIN~\cite{choi2020cain}]
			{\includegraphics[width = \patchSize, height = \patchSize]
				{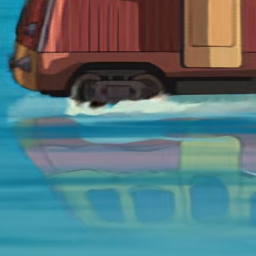}} &
			\subfloat[\scriptsize AdaCoF~\cite{Lee_2020_CVPR}]
			{\includegraphics[width = \patchSize, height = \patchSize]
				{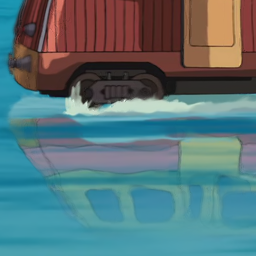}} \\ [-0.34cm] &
			\subfloat[\scriptsize SoftSplat~\cite{Niklaus_2020_CVPR}]
			{\includegraphics[width = \patchSize, height = \patchSize]
				{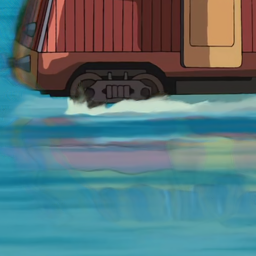}} &
			\subfloat[\scriptsize CDFI~\cite{ding2021cdfi}]
			{\includegraphics[width = \patchSize, height = \patchSize]
				{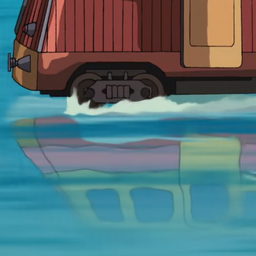}} &
			\subfloat[\scriptsize EDSC~\cite{9501506}]
			{\includegraphics[width = \patchSize, height = \patchSize]
				{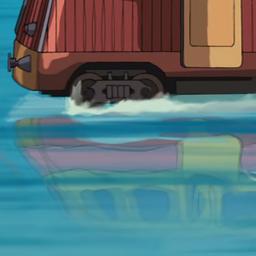}} & 
			\subfloat[\scriptsize ABME~\cite{park2021asymmetric}]
			{\includegraphics[width = \patchSize, height = \patchSize]
				{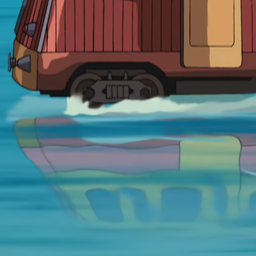}} &
			\subfloat[\scriptsize WaveletVFI]
			{\includegraphics[width = \patchSize, height = \patchSize]
				{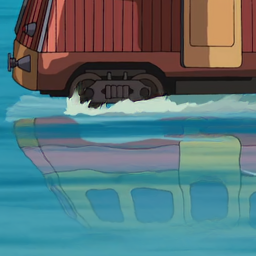}} \vspace{-1mm} \\
			
			\multirow{2}{*}[\rowArg]{
				\subfloat[\scriptsize Ground Truth Frame from ATD12K~\cite{Siyao_2021_CVPR}]
				{\includegraphics[width = \fullWidth, height = \fullHeight]
					{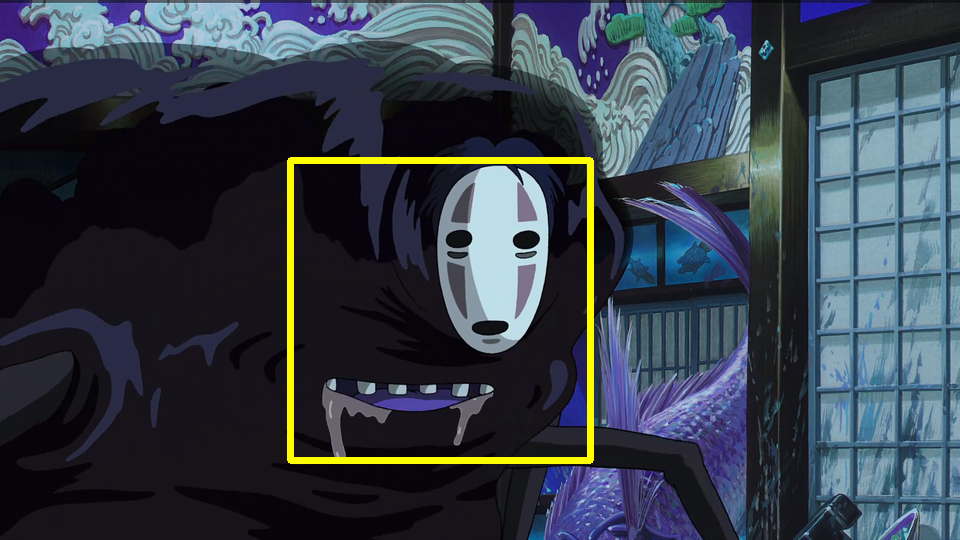}}} &
			\subfloat[\scriptsize Overlay]
			{\includegraphics[width = \patchSize, height = \patchSize]
				{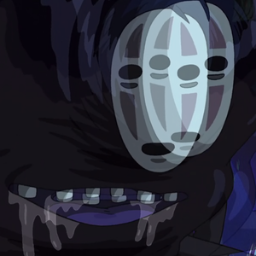}} &
			\subfloat[\scriptsize SepConv~\cite{8237299}]
			{\includegraphics[width = \patchSize, height = \patchSize]
				{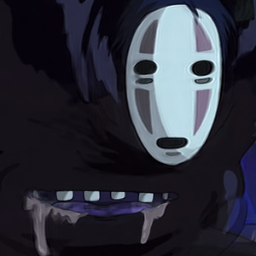}} &
			\subfloat[\scriptsize DAIN~\cite{8954114}]
			{\includegraphics[width = \patchSize, height = \patchSize]
				{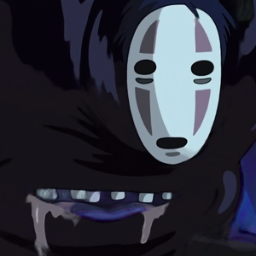}} &
			\subfloat[\scriptsize CAIN~\cite{choi2020cain}]
			{\includegraphics[width = \patchSize, height = \patchSize]
				{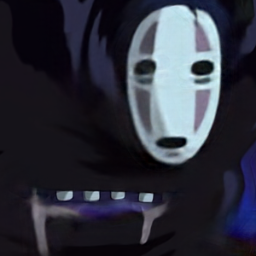}} &
			\subfloat[\scriptsize AdaCoF~\cite{Lee_2020_CVPR}]
			{\includegraphics[width = \patchSize, height = \patchSize]
				{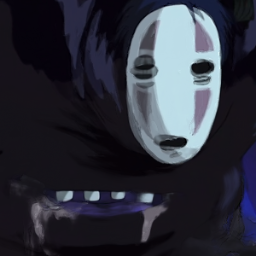}} \\ [-0.34cm] &
			\subfloat[\scriptsize SoftSplat~\cite{Niklaus_2020_CVPR}]
			{\includegraphics[width = \patchSize, height = \patchSize]
				{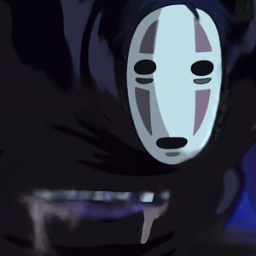}} &
			\subfloat[\scriptsize CDFI~\cite{ding2021cdfi}]
			{\includegraphics[width = \patchSize, height = \patchSize]
				{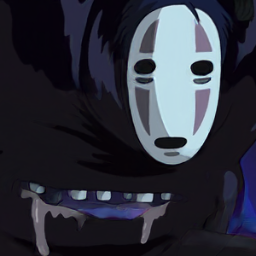}} &
			\subfloat[\scriptsize EDSC~\cite{9501506}]
			{\includegraphics[width = \patchSize, height = \patchSize]
				{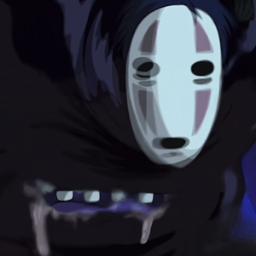}} & 
			\subfloat[\scriptsize ABME~\cite{park2021asymmetric}]
			{\includegraphics[width = \patchSize, height = \patchSize]
				{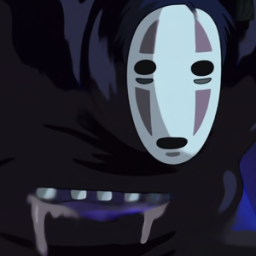}} &
			\subfloat[\scriptsize WaveletVFI]
			{\includegraphics[width = \patchSize, height = \patchSize]
				{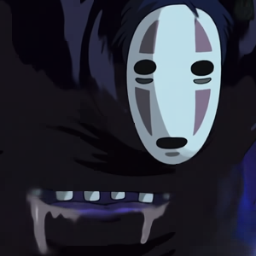}} \\
		\end{tabular}
	\end{center}
	\setlength{\abovecaptionskip}{0pt plus 2pt minus 2pt}
	\setlength{\belowcaptionskip}{0pt plus 2pt minus 2pt}
	\caption{Qualitative comparison of our WaveletVFI with other state-of-the-art frame interpolation methods on ATD12K~\cite{Siyao_2021_CVPR} dataset. Zoom in for best view.}
	\label{fig:7}
\end{figure*}

\subsubsection{Loss Functions}
For generating the target frame, we employ the same image reconstruction loss $\mathcal{L}_{r}$ as IFRNet~\cite{Kong_2022_CVPR} between the prediction $\mathrm{LL}^{0}$ and ground truth $\hat{\mathrm{LL}}^{0}$, which is the sum of two terms as
\begin{equation}
	\mathcal{L}_{r} = \rho(\mathrm{LL}^{0} - \hat{\mathrm{LL}}^{0}) + \mathcal{L}_{cen}(\mathrm{LL}^{0}, \hat{\mathrm{LL}}^{0}),
\end{equation}
where $\rho(x) = (x^2 + \epsilon^2)^{\alpha}$ with $\alpha = 0.5, \epsilon = 10^{-3}$ is the robust Charbonnier loss~\cite{413553}. $\mathcal{L}_{cen}$ is the census loss, which calculates soft Hamming distance between census-transformed image patches~\cite{Meister_2018,9882137}. Moreover, we adopt a new frequency domain reconstruction loss for better structure awareness as
\begin{equation}
	\mathcal{L}_{f} = {\sum}_{j} \rho(w_j - \hat{w}_j), 
\end{equation}
where $w_j $ and $\hat{w}_j$ are corresponding wavelet coefficient maps from the prediction set $\mathbb{W}$ and the ground truth set $\hat{\mathbb{W}}$, respectively. Finally, in order to reduce computation budget and balance different compression threshold ratio selection, we propose a computation cost regularization term as
\begin{equation}
	\mathcal{L}_{c} = {\sum}_{k=1}^{m} h_k \cdot \mathcal{C}(\mathcal{N}_{\mathrm{WS}}(\cdots; h_k)) \, / \, (H \times W), 
\end{equation}
where $\mathcal{C}$ is the FLOPs counter, $H$ and $W$ represent the height and width of original input resolution. Our final objective function combines above three components with weighting parameters of $\alpha$ and $\beta$, where $\beta$ controls the trade-off between accuracy and efficiency, that is formulated as
\begin{equation}
	\mathcal{L} = \mathcal{L}_{r} + \alpha \mathcal{L}_{f} + \beta \mathcal{L}_{c}.
	\label{eq:15}
\end{equation}

\section{Experiments} \label{Experiments}
In this section, we first introduce the datasets used for training and test, and implementation details about the learning strategy. Then, we compare the proposed framework with recent state-of-the-art VFI methods on the commonly used low resolution, high resolution and animation frame interpolation benchmarks quantitatively and qualitatively. Finally, we carry out ablation study for analysis, and do more discussion.

\subsection{Datasets}
In this work, we supervise the proposed WaveletVFI on the training split of Vimeo90K~\cite{xue2019video}, and test it on multiple datasets summarized as follows: \textbf{1) Vimeo90K~\cite{xue2019video}} is a widely-used dataset for video processing tasks. There are 3,782 triplets with 448 $\times$ 256 resolution in the test set. \textbf{2) ATD12K~\cite{Siyao_2021_CVPR}} is an animation frame interpolation benchmark, where there are 2,000 triplets from diverse cartoon scenarios in test datasets. Note that we adopt the 960 $\times$ 540 resolution part to cover diverse video resolutions for more sufficient evaluation, which is different from the 1080p test part reported in the original paper. \textbf{3) Xiph~\cite{Montgomery1994Xiph}} contains 30 raw video sequences that is originally used for testing video codecs. For frame interpolation, we follow the data processing operations in SoftSplat~\cite{Niklaus_2020_CVPR} to generate Xiph-2K test set by down sampling 4K videos and generate Xiph-4K test set by center cropping 2K patches. There are 392 frame triplets of resolution 2048 $\times$ 1080 in both Xiph-2K and Xiph-4K benchmarks.

\begin{table*}[t]
	\centering
	\renewcommand{\arraystretch}{1.0}
	{\scriptsize
		\setlength\tabcolsep{4.0pt}
		\caption{Quantitative comparison with recent state-of-the-art frame interpolation methods on Vimeo90K, ATD12K, Xiph-2K and Xiph-4K benchmarks. Computation complexity is measured in tera-FLOPs (TFLOPs). For each item, \\the best result is \textcolor{red}{\textbf{boldfaced}}, and the second best is \textcolor{blue}{\underline{underlined}}.}
		\resizebox{0.98\textwidth}{!}{
			\begin{tabular}{lccccccccccccc}
				\toprule
				\multirow{2}[1]{*}{Method} & Params & \multicolumn{3}{c}{Vimeo90K} & \multicolumn{3}{c}{ATD12K} & \multicolumn{3}{c}{Xiph-2K} & \multicolumn{3}{c}{Xiph-4K} \\
				\cmidrule(lr){3-5} \cmidrule(lr){6-8} \cmidrule(lr){9-11} \cmidrule(lr){12-14}
				& (M) & TFLOPs & PSNR & SSIM & TFLOPs & PSNR & SSIM & TFLOPs & PSNR & SSIM & TFLOPs & PSNR & SSIM \\
				\midrule
				SepConv~\cite{8237299} & 21.7 & 0.108 & 33.79 & 0.970 & 0.487 & 27.40 & 0.950 & 2.078 & 34.77 & 0.929 & 2.078 & 32.06 & 0.880 \\
				DAIN~\cite{8954114} & 24.0 & 0.686 & 34.71 & 0.976 & 3.099 & 27.38 & 0.955 & 13.22 & 35.97 & 0.940 & 13.22 & 33.51 & 0.898 \\
				CAIN~\cite{choi2020cain} & 42.8 & 0.162 & 34.65 & 0.973 & 0.734 & 25.28 & 0.952 & 3.133 & 35.21 & 0.937 & 3.133 & 32.56 & 0.901 \\
				AdaCoF+~\cite{Lee_2020_CVPR} & 22.9 & 0.282 & 34.56 & 0.959 & 1.273 & 27.39 & 0.937 & 5.433 & 35.09 & 0.931 & 5.433 & 32.19 & 0.882 \\
				SoftSplat~\cite{Niklaus_2020_CVPR} & 12.2 & 0.112 & 36.10 & 0.970 & 0.506 & 28.22 & \textcolor{blue}{\underline{0.957}} & 2.160 & \textcolor{blue}{\underline{36.62}} & 0.944 & 2.160 & \textcolor{blue}{\underline{33.60}} & 0.901 \\
				BMBC~\cite{BMBC} & \textcolor{blue}{\underline{11.0}} & 0.311 & 35.06 & 0.964 & 1.405 & 27.68 & 0.945 & 5.994 & 32.82 & 0.928 & 5.994 & 31.19 & 0.880 \\
				CDFI~\cite{ding2021cdfi} & \textcolor{red}{\bf 5.0} & 0.102 & 35.17 & 0.964 & 0.463 & 28.15 & 0.950 & 1.977 & 35.50 & 0.960 & 1.977 & 32.50 & 0.932 \\
				ABME~\cite{park2021asymmetric} & 18.1 & 0.161 & \textcolor{blue}{\underline{36.18}} & \textcolor{red}{\bf 0.981} & 0.728 & 28.71 & \textcolor{red}{\bf 0.959} & 3.108 & 35.18 & 0.964 & 3.108 & 32.36 & 0.940 \\
				CAIN-SD~\cite{Choi_2021_ICCV} & $>$ 42 & - & - & - & - & - & - & \textcolor{blue}{\underline{1.598}} & 34.68 & 0.924 & 1.983 & 32.92 & 0.893 \\
				IFRNet-L~\cite{Kong_2022_CVPR} & 19.7 & \textcolor{blue}{\underline{0.098}} & \textcolor{red}{\bf 36.20} & \textcolor{red}{\bf 0.981} & \textcolor{blue}{\underline{0.444}} & \textcolor{blue}{\underline{28.78}} & 0.956 & 1.896 & \textcolor{red}{\bf 36.63} & \textcolor{red}{\bf 0.966} & \textcolor{blue}{\underline{1.896}} & 33.58 & \textcolor{blue}{\underline{0.944}} \\
				\midrule
				WaveletVFI (Ours) & 19.4 & \textcolor{red}{\bf 0.081} & 35.58 & \textcolor{blue}{\underline{0.978}} & \textcolor{red}{\bf 0.274} & \textcolor{red}{\bf 28.79} & 0.956 & \textcolor{red}{\bf 1.480} & 36.32 & \textcolor{blue}{\underline{0.965}} & \textcolor{red}{\bf 1.428} & \textcolor{red}{\bf 33.61} & \textcolor{red}{\bf 0.945} \\
				\bottomrule
		\end{tabular}}
		\label{tab:1}}
\end{table*}

\begin{table}[t]
	\centering
	\renewcommand{\arraystretch}{1.1}
	{\scriptsize
		\setlength\tabcolsep{6.0pt}
		\caption{Comparison of running time and memory usage on Xiph-4K. Time and memory are measured on one Tesla V100 \\ GPU under PyTorch implementation.}
		\resizebox{0.48\textwidth}{!}{
			\begin{tabular}{c|c|c|c|c}
				\toprule
				Method & DAIN & CAIN & AdaCoF+ & SoftSplat \\
				\hline
				Time (s) & 2.39 & 0.17 & 0.32 & 0.41 \\
				\hline
				Memory (GB) & 15.9 & 4.7 & 12.1 & 8.8 \\
				\midrule
				Method & BMBC & CDFI & ABME & WaveletVFI \\
				\hline
				Time (s) & 9.10 & 0.92 & 1.63 & 0.19 \\
				\hline
				Memory (GB) & 27.2 & 27.9 & 17.2 & 7.0 \\
				\bottomrule
		\end{tabular}}
		\label{tab:2}}
\end{table}

\subsection{Implementation Details} \label{Implementation Details}
We implement the proposed WaveletVFI in PyTorch and adopt a two step learning schedule to train our algorithm on Vimeo90K training set from scratch. First, we train the $\mathcal{N}_{\mathrm{MP}}$ and $\mathcal{N}_{\mathrm{WS}}$ but without the threshold classifier for 300 epochs as initialization, where the compression threshold ratio $\eta$ is set to 0, weighting parameters $\alpha$ and $\beta$ in Eq.~\ref{eq:15} are set to 0.01 and 0 respectively. Then, we load the pre-trained parameters in step 1 and fine-tune the whole WaveletVFI framework with proposed dynamic threshold ratio selection approach for another 100 epochs to learn instance-aware threshold ratio selection, that considers the trade-off between accuracy and efficiency. In this stage, we set $\alpha$ and $\beta$ to be 0.01 and 1 separately. All parameters that need to update are optimized by AdamW~\cite{Loshchilov_2019} algorithm, and the model is trained with total batch size 24 on four NVIDIA Tesla V100 GPUs. In both steps, the learning rate is initially set to $1 \times 10^{-4}$, and gradually decays to $1 \times 10^{-5}$ following a cosine attenuation schedule. During training, we augment the triplet samples by random horizontal and vertical flipping, rotating, reversing sequence order and random cropping patches with size 256 $\times$ 256. Following the common practice of $t=0.5$, all compared approaches only interpolate one middle frame in the experiments.

\subsection{Comparison with the State-of-the-Arts}
We compare proposed WaveletVFI with state-of-the-art VFI methods, including kernel-based SepConv~\cite{8237299}, CAIN~\cite{choi2020cain}, AdaCoF~\cite{Lee_2020_CVPR}, CDFI~\cite{ding2021cdfi} and CAIN-SD~\cite{Choi_2021_ICCV}, flow-based DAIN~\cite{8954114}, SoftSplat~\cite{Niklaus_2020_CVPR}, BMBC~\cite{BMBC}, ABME~\cite{park2021asymmetric} and IFRNet~\cite{Kong_2022_CVPR}. Common metrics, such as Peak Signal-to-Noise Ratio (PSNR) and Structural Similarity (SSIM)~\cite{1284395} are adopted for quantitative evaluation. For the computation complexity, we calculate the number of floating-point operations (FLOPs) and average per sample FLOPs over specific dataset.

\subsubsection{Quantitative Comparison}
As shown in Table~\ref{tab:1}, proposed approach always requires the smallest FLOPs than others, while achieving better or comparable accuracy. On the ATD12K~\cite{Siyao_2021_CVPR} 540p animation benchmark, our method takes only 59\% computation cost of the efficient CDFI~\cite{ding2021cdfi}, while obtaining 0.64 dB performance improvement. The method IFRNet~\cite{Kong_2022_CVPR} and ABME~\cite{park2021asymmetric} also achieve similar accuracy as ours, however, proposed WaveletVFI only uses 62\% or 38\% multiply-add operations respectively with comparable model size. As for the high resolution Xiph~\cite{Montgomery1994Xiph} datasets, our approach performs best in both PSNR and SSIM on the more challenging Xiph-4K benchmark, while only falls behind IFRNet~\cite{Kong_2022_CVPR} and SoftSplat~\cite{Niklaus_2020_CVPR} on the Xiph-2K test set. Similar as ours, SoftSplat is also a two-stage flow-based frame interpolation method. However, thanks to the efficiency of proposed dynamic compression threshold selection approach in wavelet domain, we can achieve better or on par accuracy than SoftSplat, while requiring only 67\% computation on the high resolution Xiph benchmarks. Compared with CAIN-SD~\cite{Choi_2021_ICCV}, that is also a dynamic VFI method for reducing computation complexity, we achieve smaller FLOPs while obtaining significant accuracy gain, \textit{i.e.}, 1.64 dB better on Xiph-2K and 0.69 dB better on Xiph-4K. In regard to real deployment, as shown in Table~\ref{tab:2}, we test inference time and peak GPU memory usage of several well-behaved VFI methods on one Tesla V100 GPU under PyTorch on Xiph-4K. It can be seen that proposed approach only falls behind CAIN~\cite{choi2020cain} but outperforms others. Note that our sparse convolution is simulated by traditional convolution multiplied with sparse mask which supports end-to-end optimization in training. During inference, we follow the approach in WaveletMonodepth~\cite{Ramamonjisoa_2021_CVPR}, and can replace it by more efficient operations.

In summary, considering 8 accuracy metrics, including PSNR and SSIM on Vimeo90K, ATD12K, Xiph-2K and Xiph-4K datasets, our approach ranks 1st 3 times, 2nd 1 times, and 3rd 3 times. Moreover, we always consume the least amount of computation. Above quantitative results have demonstrated the good comprehensive performance of our approaches.

\begin{figure*}[t]
	\captionsetup[subfloat]{labelformat=empty,captionskip=-1pt}
	\begin{center}
		\newcommand{\rowArg}{2.11cm}
		\newcommand{\fullHeight}{4.27cm}
		\newcommand{\fullWidth}{7.59cm}
		\newcommand{\patchSize}{1.96cm}
		\scriptsize
		\setlength\tabcolsep{0.05cm}
		\begin{tabular}[b]{c c c c c c}
			\multirow{2}{*}[\rowArg]{
				\subfloat[\scriptsize Ground Truth Frame from Xiph-4K~\cite{Montgomery1994Xiph}]
				{\includegraphics[width = \fullWidth, height = \fullHeight]
					{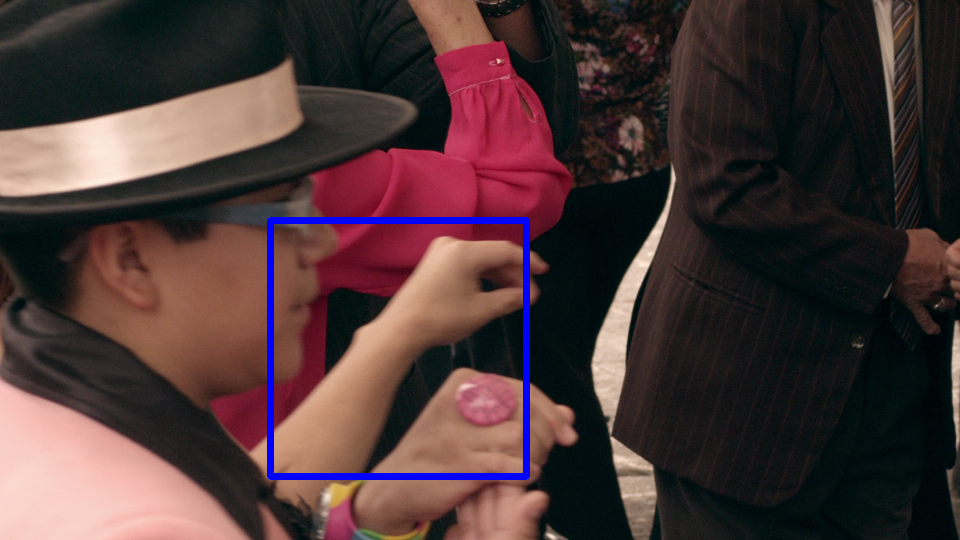}}} &
			\subfloat[\scriptsize Overlay]
			{\includegraphics[width = \patchSize, height = \patchSize]
				{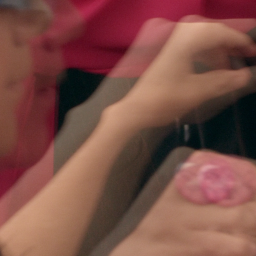}} &
			\subfloat[\scriptsize SepConv~\cite{8237299}]
			{\includegraphics[width = \patchSize, height = \patchSize]
				{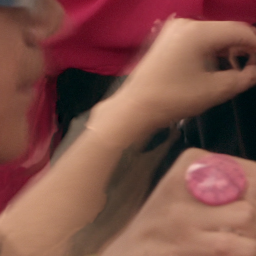}} &
			\subfloat[\scriptsize DAIN~\cite{8954114}]
			{\includegraphics[width = \patchSize, height = \patchSize]
				{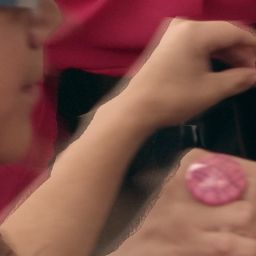}} &
			\subfloat[\scriptsize CAIN~\cite{choi2020cain}]
			{\includegraphics[width = \patchSize, height = \patchSize]
				{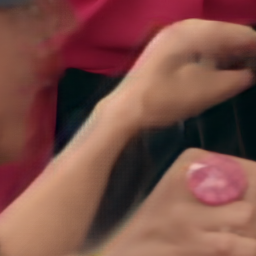}} &
			\subfloat[\scriptsize AdaCoF~\cite{Lee_2020_CVPR}]
			{\includegraphics[width = \patchSize, height = \patchSize]
				{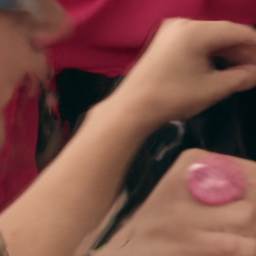}} \\ [-0.34cm] &
			\subfloat[\scriptsize SoftSplat~\cite{Niklaus_2020_CVPR}]
			{\includegraphics[width = \patchSize, height = \patchSize]
				{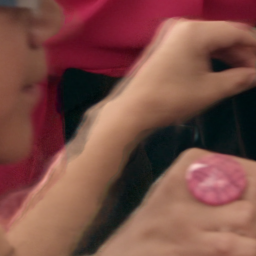}} &
			\subfloat[\scriptsize CDFI~\cite{ding2021cdfi}]
			{\includegraphics[width = \patchSize, height = \patchSize]
				{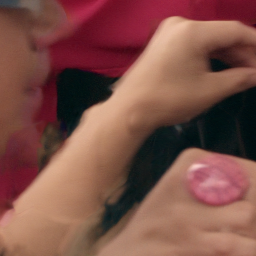}} &
			\subfloat[\scriptsize EDSC~\cite{9501506}]
			{\includegraphics[width = \patchSize, height = \patchSize]
				{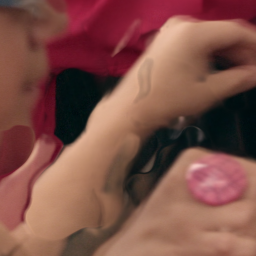}} & 
			\subfloat[\scriptsize ABME~\cite{park2021asymmetric}]
			{\includegraphics[width = \patchSize, height = \patchSize]
				{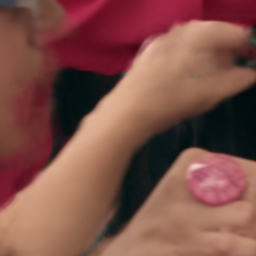}} &
			\subfloat[\scriptsize WaveletVFI]
			{\includegraphics[width = \patchSize, height = \patchSize]
				{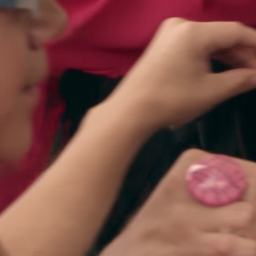}} \vspace{-1mm} \\
			
			\multirow{2}{*}[\rowArg]{
				\subfloat[\scriptsize Ground Truth Frame from Xiph-4K~\cite{Montgomery1994Xiph}]
				{\includegraphics[width = \fullWidth, height = \fullHeight]
					{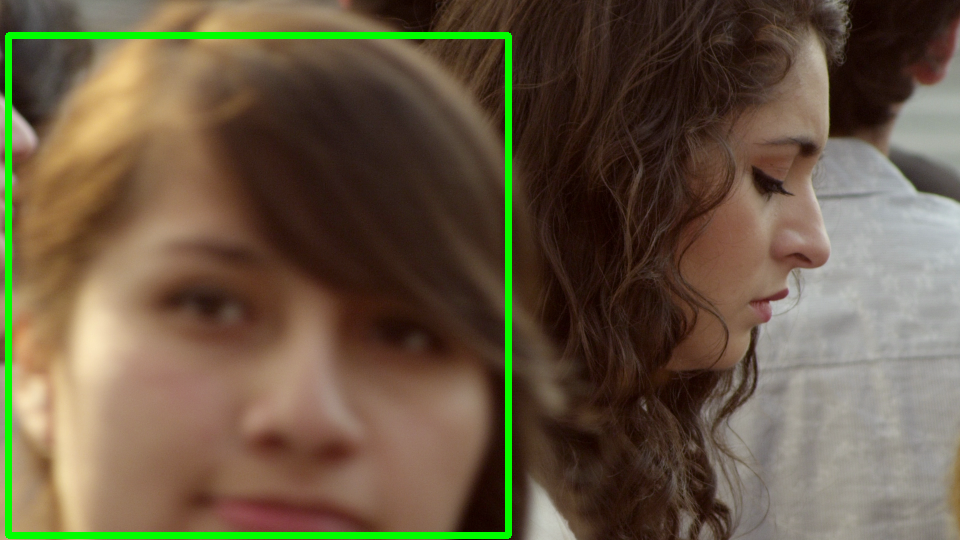}}} &
			\subfloat[\scriptsize Overlay]
			{\includegraphics[width = \patchSize, height = \patchSize]
				{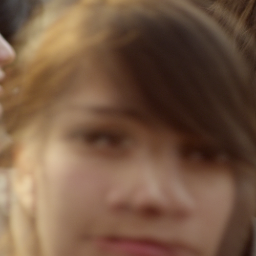}} &
			\subfloat[\scriptsize SepConv~\cite{8237299}]
			{\includegraphics[width = \patchSize, height = \patchSize]
				{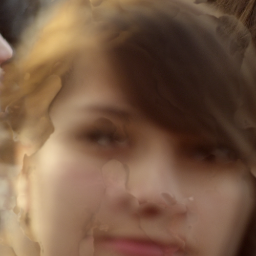}} &
			\subfloat[\scriptsize DAIN~\cite{8954114}]
			{\includegraphics[width = \patchSize, height = \patchSize]
				{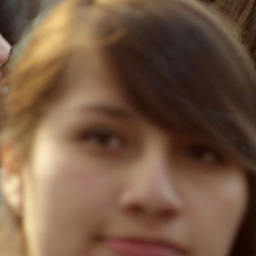}} &
			\subfloat[\scriptsize CAIN~\cite{choi2020cain}]
			{\includegraphics[width = \patchSize, height = \patchSize]
				{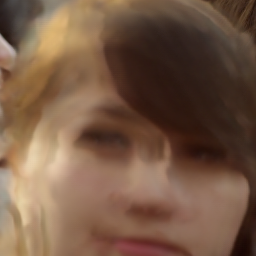}} &
			\subfloat[\scriptsize AdaCoF~\cite{Lee_2020_CVPR}]
			{\includegraphics[width = \patchSize, height = \patchSize]
				{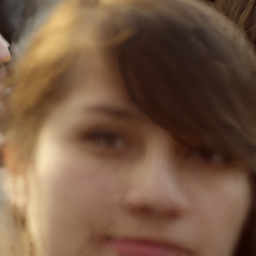}} \\ [-0.34cm] &
			\subfloat[\scriptsize SoftSplat~\cite{Niklaus_2020_CVPR}]
			{\includegraphics[width = \patchSize, height = \patchSize]
				{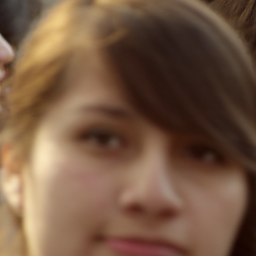}} &
			\subfloat[\scriptsize CDFI~\cite{ding2021cdfi}]
			{\includegraphics[width = \patchSize, height = \patchSize]
				{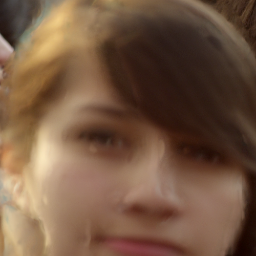}} &
			\subfloat[\scriptsize EDSC~\cite{9501506}]
			{\includegraphics[width = \patchSize, height = \patchSize]
				{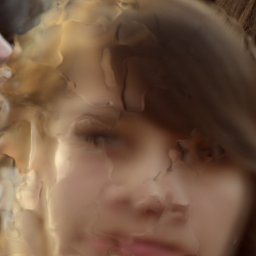}} & 
			\subfloat[\scriptsize ABME~\cite{park2021asymmetric}]
			{\includegraphics[width = \patchSize, height = \patchSize]
				{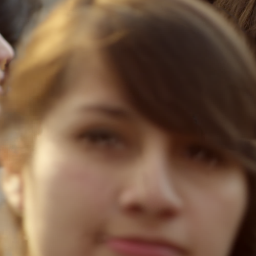}} &
			\subfloat[\scriptsize WaveletVFI]
			{\includegraphics[width = \patchSize, height = \patchSize]
				{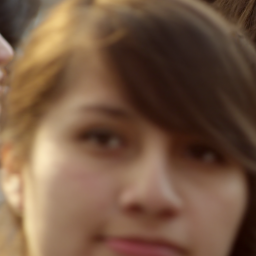}} \\
		\end{tabular}
	\end{center}
	\setlength{\abovecaptionskip}{0pt plus 2pt minus 2pt}
	\setlength{\belowcaptionskip}{0pt plus 2pt minus 2pt}
	\caption{Qualitative comparison of our WaveletVFI with other state-of-the-art frame interpolation methods on Xiph-4K~\cite{Montgomery1994Xiph} dataset. Zoom in for best view.}
	\label{fig:8}
\end{figure*}

\begin{table}[t]
	\centering
	\renewcommand{\arraystretch}{1.0}
	\caption{Ablation of complementary context encoders $\mathcal{E}_{\mathrm{WS1}}, \mathcal{E}_{\mathrm{WS2}}$, frequency reconstruction loss $\mathcal{L}_{f}$ and wavelet domain interpolation $W$ for $\mathcal{N}_{\mathrm{WS}}$ during the first training step.}
	{\footnotesize
		\setlength\tabcolsep{9.0pt}
		\resizebox{0.48\textwidth}{!}{
			\begin{tabular}{c|cccc|cc}
				\toprule
				\multirow{2}{*}{ID} & \multirow{2}{*}{$\mathcal{E}_{\mathrm{WS1}}$} & \multirow{2}{*}{$\mathcal{E}_{\mathrm{WS2}}$} & \multirow{2}{*}{$\mathcal{L}_{f}$} & \multirow{2}{*}{$W$} & \multicolumn{2}{c}{Vimeo90K} \\
				& & & & & \multicolumn{1}{c}{PSNR} & \multicolumn{1}{c}{SSIM} \\
				\midrule
				E1 & \cmark & \xmark & \xmark & \cmark & 34.58 & 0.965 \\
				E2 & \xmark & \cmark & \xmark & \cmark & 35.33 & 0.972 \\
				E3 & \cmark & \cmark & \xmark & \cmark & 35.56 & 0.976 \\
				E4 & \cmark & \cmark & \cmark & \cmark & \textbf{35.60} & \textbf{0.979} \\
				E5 & \cmark & \cmark & \xmark & \xmark & 35.56 & 0.974 \\
				\bottomrule
	\end{tabular}}}
	\label{tab:3}
\end{table}

\subsubsection{Qualitative Comparison}
To visually compare with other SOTA methods, we show two examples from ATD12K and Xiph-4K in Fig.~\ref{fig:7} and Fig.~\ref{fig:8}, respectively. The first example in Fig.~\ref{fig:7} depicts a bus driving on the water, where proposed approach can synthesize the reflection of this cartoon bus realistically, while predictions from other methods behave blurry and twisty. The second example in Fig.~\ref{fig:7} shows the character of No-Face in Spirited Away, where our method can generate clearer white face and more reasonable teeth. In Fig.~\ref{fig:8}, the first example shows a fast moving man waving his arm, which is a challenging case in Xiph-4K. It is obvious that proposed WaveletVFI can generate sharp motion boundary, while results of other methods contain ghosting artifacts. As for the second example in Fig.~\ref{fig:8}, interpolated frame of our approach looks more faithful and distinct.

\subsection{Ablation Study}
In this part, we analyze proposed contributions in network structure, loss function and hyper-parameters to explore the diverse characteristics of frame interpolation in wavelet domain and verify the effectiveness of proposed approaches.

\subsubsection{Complementary Encoder, Frequency Loss and Wavelet Domain Interpolation}
As shown in Table~\ref{tab:3}, we carry out ablation to verify the effectiveness of complementary context encoders $\mathcal{E}_{\mathrm{WS1}}, \mathcal{E}_{\mathrm{WS2}}$, frequency reconstruction loss $\mathcal{L}_{f}$ and wavelet domain frame interpolation $W$. We selectively remove $\mathcal{E}_{\mathrm{WS1}}$ or $\mathcal{E}_{\mathrm{WS2}}$ and enlarge feature channels of the remaining context encoder to be the same as original one for fair comparison. As listed in the first three rows of Table~\ref{tab:3}, $\mathcal{E}_{\mathrm{WS2}}$ behaves more important than $\mathcal{E}_{\mathrm{WS1}}$. It is because that $\mathcal{E}_{\mathrm{WS2}}$ jointly models scene texture, motion and occlusion, while $\mathcal{E}_{\mathrm{WS1}}$ is more focused on original contextual details. The combination of $\mathcal{E}_{\mathrm{WS1}}$ and $\mathcal{E}_{\mathrm{WS2}}$ achieves best results, demonstrating they are mutually benefit. Moreover, as listed of E4 in Table~\ref{tab:3}, we can obtain better performance than E3 by introducing an additional frequency reconstruction loss $\mathcal{L}_{f}$. In this setting, improvement of SSIM is more obvious, verifying $\mathcal{L}_{f}$ can help generate better scene structure. The last experiment E5 uses the same synthesis network structure but directly predicts target frame in color space. It concludes that frame interpolation in wavelet domain behaves a little better than in color space, verifying the rationality of progressive IDWT based target frame reconstruction for VFI.

\begin{table*}[t]
	\centering
	\renewcommand{\arraystretch}{1.0}
	{\scriptsize
		\setlength\tabcolsep{4.4pt}
		\caption{Ablation study of different fixed compression threshold ratio $\eta$ for accuracy vs efficiency trade-off on multiple datasets.}
		\resizebox{0.98\textwidth}{!}{
			\begin{tabular}{cccccccccccccc}
				\toprule
				\multirow{2}[1]{*}{Dynamic} & \multirow{2}[1]{*}{$\eta$} & \multicolumn{3}{c}{Vimeo90K} & \multicolumn{3}{c}{ATD12K} & \multicolumn{3}{c}{Xiph-2K} & \multicolumn{3}{c}{Xiph-4K} \\
				\cmidrule(lr){3-5} \cmidrule(lr){6-8} \cmidrule(lr){9-11} \cmidrule(lr){12-14}
				& & TFLOPs & PSNR & SSIM & TFLOPs & PSNR & SSIM & TFLOPs & PSNR & SSIM & TFLOPs & PSNR & SSIM \\
				\midrule
				\xmark & 0.0000 & 0.090 & 35.71 & 0.9791 & 0.409 & 28.83 & 0.9566 & 1.746 & 36.40 & 0.9664 & 1.746 & 33.62 & 0.9453 \\
				\xmark & 0.0025 & 0.087 & 35.70 & 0.9789 & 0.346 & 28.83 & 0.9565 & 1.660 & 36.39 & 0.9662 & 1.666 & 33.61 & 0.9452 \\
				\xmark & 0.0050 & 0.084 & 35.63 & 0.9780 & 0.314 & 28.82 & 0.9563 & 1.569 & 36.32 & 0.9646 & 1.558 & 33.58 & 0.9449 \\
				\xmark & 0.0075 & 0.081 & 35.49 & 0.9762 & 0.289 & 28.80 & 0.9558 & 1.491 & 36.16 & 0.9620 & 1.465 & 33.51 & 0.9441 \\
				\xmark & 0.0100 & 0.079 & 35.30 & 0.9738 & 0.270 & 28.78 & 0.9552 & 1.420 & 35.94 & 0.9587 & 1.385 & 33.43 & 0.9428 \\
				\xmark & 0.0125 & 0.077 & 35.05 & 0.9709 & 0.256 & 28.76 & 0.9546 & 1.356 & 35.68 & 0.9557 & 1.316 & 33.32 & 0.9413 \\
				\xmark & 0.0150 & 0.075 & 34.76 & 0.9675 & 0.245 & 28.73 & 0.9538 & 1.301 & 35.38 & 0.9528 & 1.256 & 33.19 & 0.9396 \\
				\bottomrule
		\end{tabular}}
		\label{tab:4}}
\end{table*}

\begin{figure}[t]
	\includegraphics[width=0.98\linewidth]{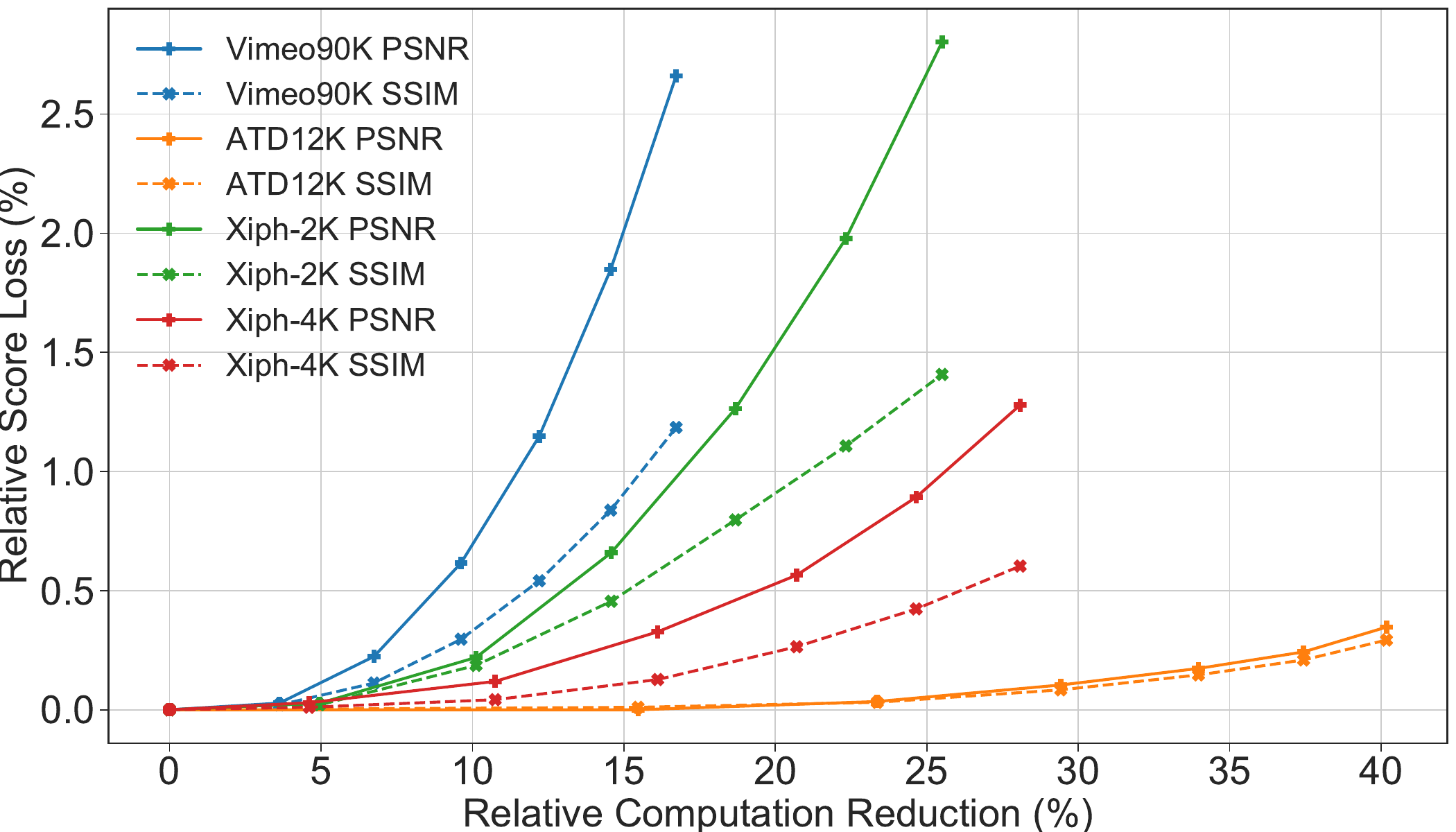}
	\caption{Analysis of accuracy vs efficiency by fixed compression threshold $\eta$.}
	\label{fig:9}
\end{figure}

\begin{figure}[t]
	\includegraphics[width=0.98\linewidth]{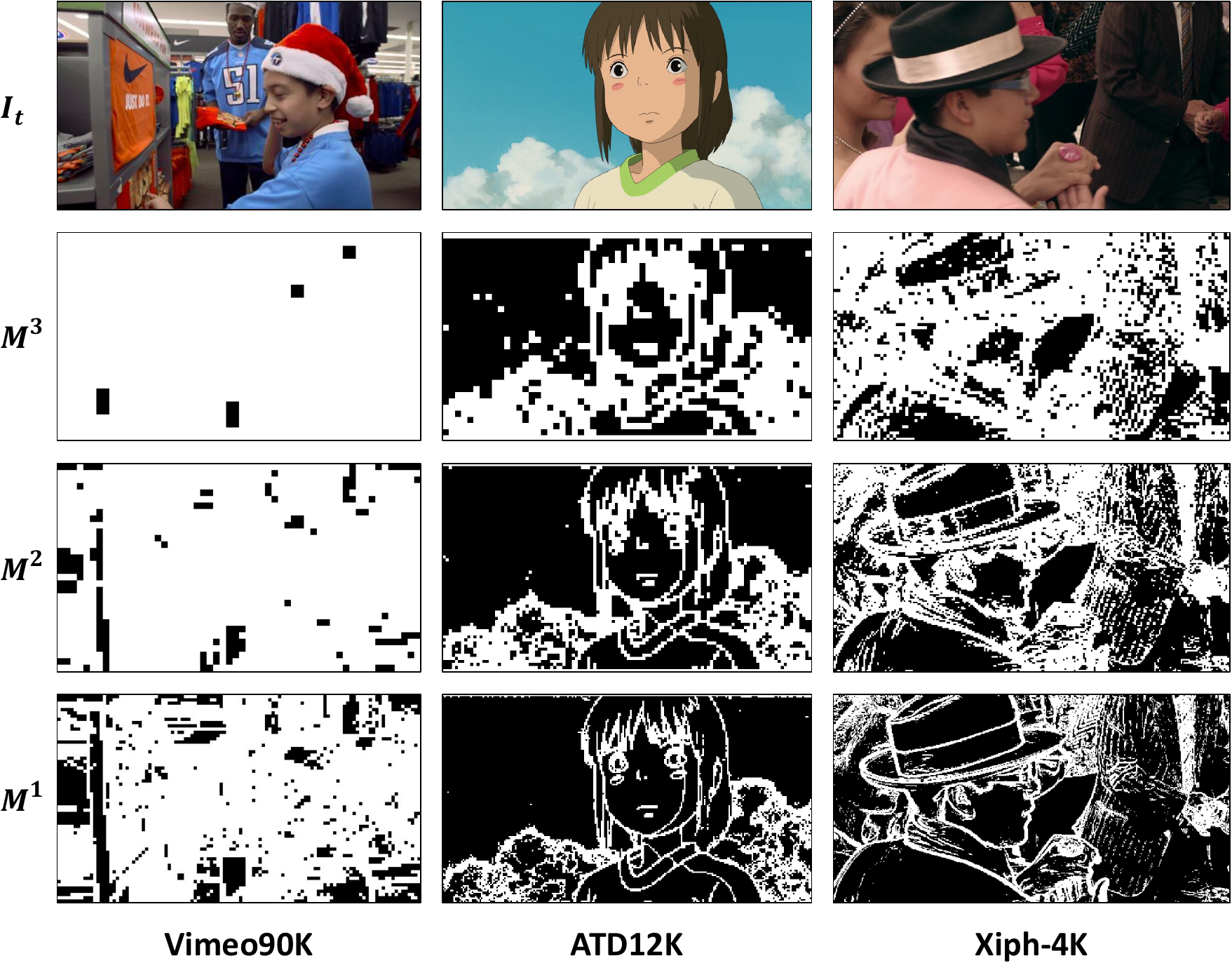}
	\caption{Predicted target frame and sparse valid masks on diverse datasets.}
	\label{fig:10}
\end{figure}

\begin{table}[t]
	\centering
	\renewcommand{\arraystretch}{1.1}
	{\scriptsize
		\setlength\tabcolsep{2.8pt}
		\caption{Motion magnitude statistics on multiple datasets.}
		\resizebox{0.46\textwidth}{!}{
			\begin{tabular}{c|cccc}
				\toprule
				Dataset & Vimeo90K & ATD12K & Xiph-2K & Xiph-4K \\
				\midrule
				Mean Value & 6.06 & 19.4 & 10.9 & 25.3 \\
				Standard Deviation & 6.12 & 32.3 & 12.5 & 33.3 \\
				\bottomrule
		\end{tabular}}
		\label{tab:5}}
\end{table}

\begin{figure}[t]
	\centering	
	\includegraphics[width=0.48\linewidth]{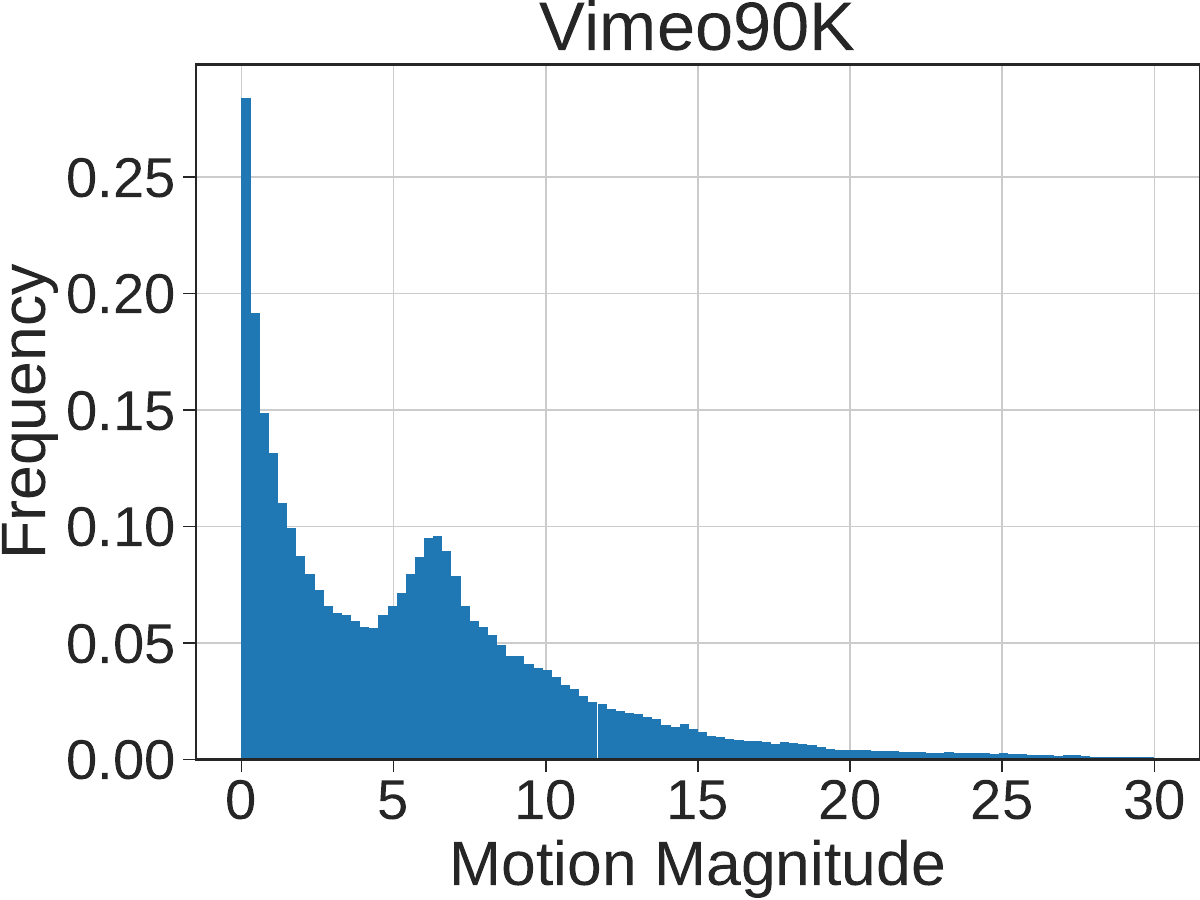}
	\includegraphics[width=0.48\linewidth]{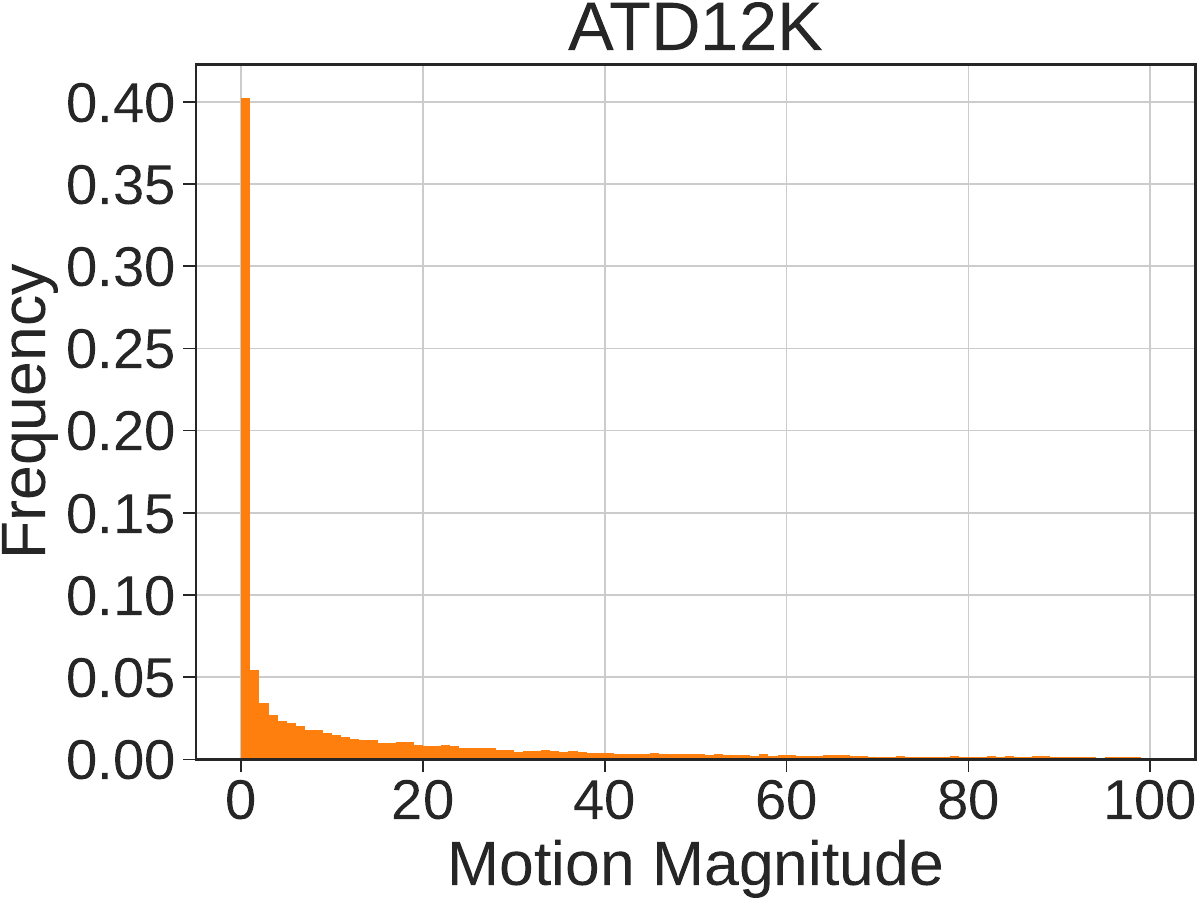}
	\vspace{2mm} \\
	\includegraphics[width=0.48\linewidth]{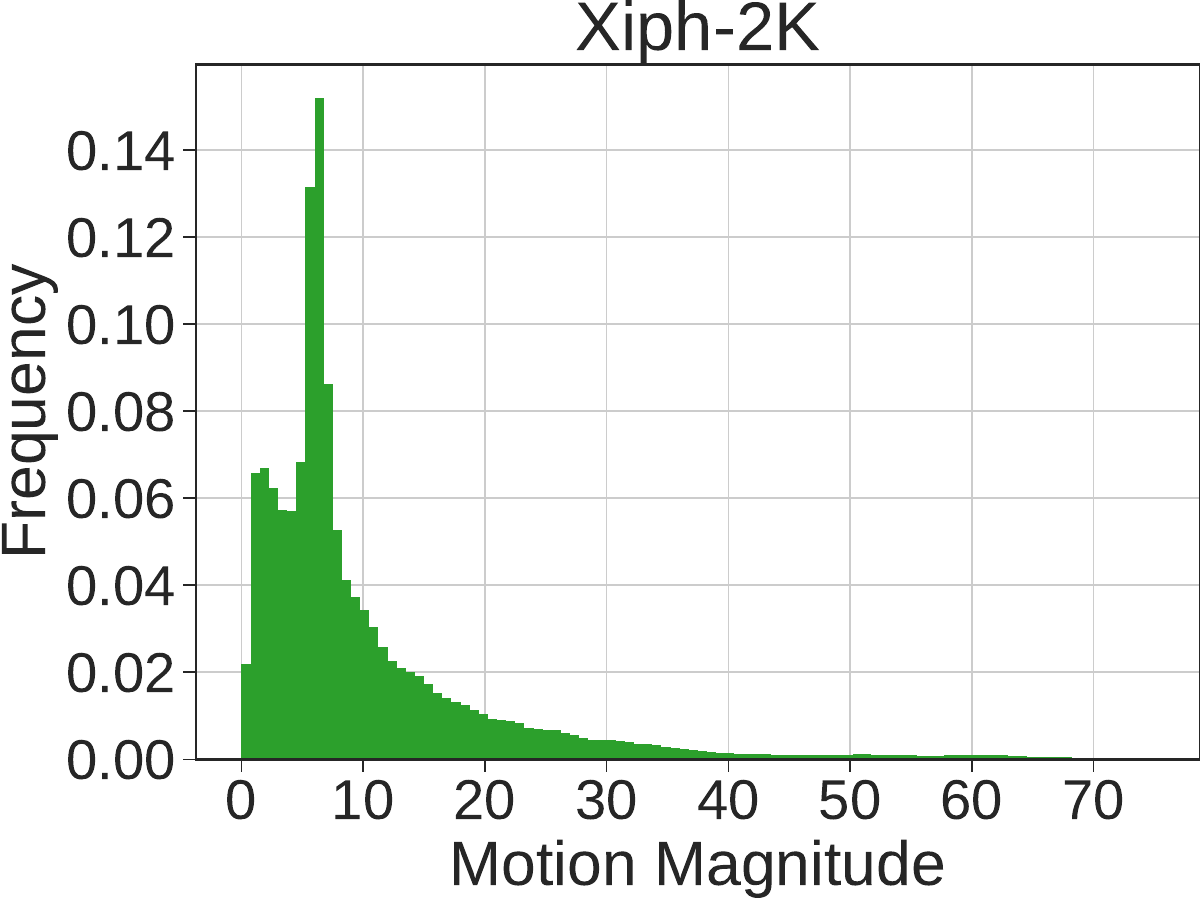}
	\includegraphics[width=0.48\linewidth]{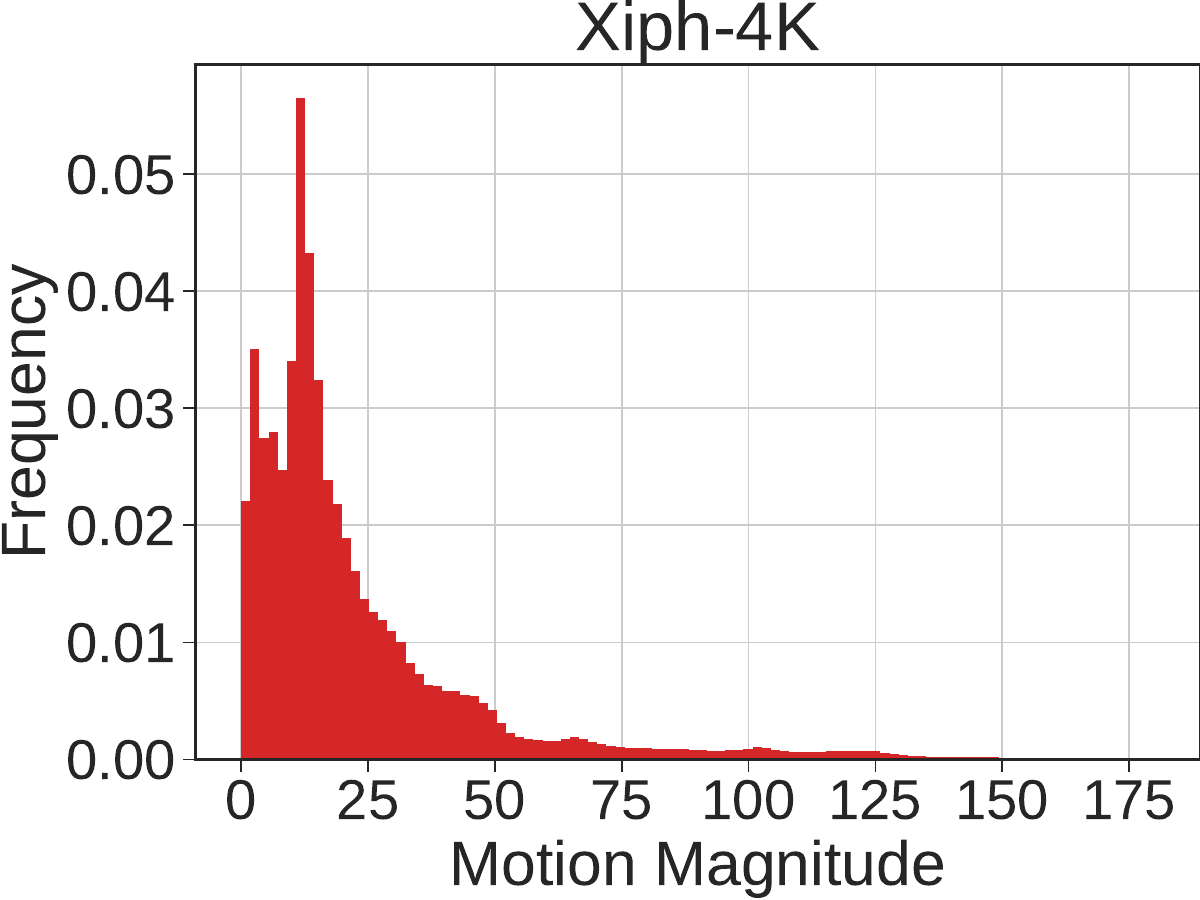}
	\caption{Motion magnitude statistics on multiple datasets.}
	\label{fig:11}
\end{figure}

\subsubsection{Accuracy vs Efficiency Trade-off by Wavelet Sparsity}
To demonstrate the superiority of sparse representation in wavelet domain for efficient frame interpolation, we explore the accuracy vs efficiency relationship by setting different fixed compression threshold ratio $\eta$ in the second training step, and the corresponding $\eta$ is also used during evaluation. We select 7 typical $\eta$ values of $0.0000, 0.0025, 0.0050, \dots, 0.0150$ to carry out above experiments on diverse datasets, whose results are summarized in Table~\ref{tab:4}. For better intuitive understanding, we depict the relative change curves of score loss ratio against computation reduction ratio in Fig.~\ref{fig:9}, where $\beta$ is set to 0 in all these cases. In Fig.~\ref{fig:9}, each point stands for an experiment result of one specific $\eta$ value, and $\eta$ gradually increases from left to right in a specific line. As is expected, larger $\eta$ will result in less computation and lower performance, however, the relative change rate of accuracy vs efficiency shows big difference among different datasets. On Vimeo90K, PSNR drops about 0.7\% when computation is reduced by 10\%. While on ATD12K, PSNR only drops about 0.4\% even computation is reduced by 40\%. It concludes that for a fixed $\eta$, relative computation reduction is more obvious when there is more sparse low frequency texture in this dataset. Fig.~\ref{fig:10} visually supports this conclusion by showing predicted multi-scale sparse valid masks $M^{l}, l \in \{1, 2, 3\}$ on different VFI datasets under the same compression threshold ratio $\eta=0.01$. This phenomenon also appears in traditional image compression~\cite{10.1007/978-3-030-58565-5_19,859238,862633}, which means that high resolution or cartoon images are more spatially redundant to achieve higher compression ratio under the same compressed image quality.

\subsubsection{Dynamic Compression Threshold Ratio Selection}
\begin{figure*}[t]
	\centering	
	\includegraphics[width=0.245\linewidth]{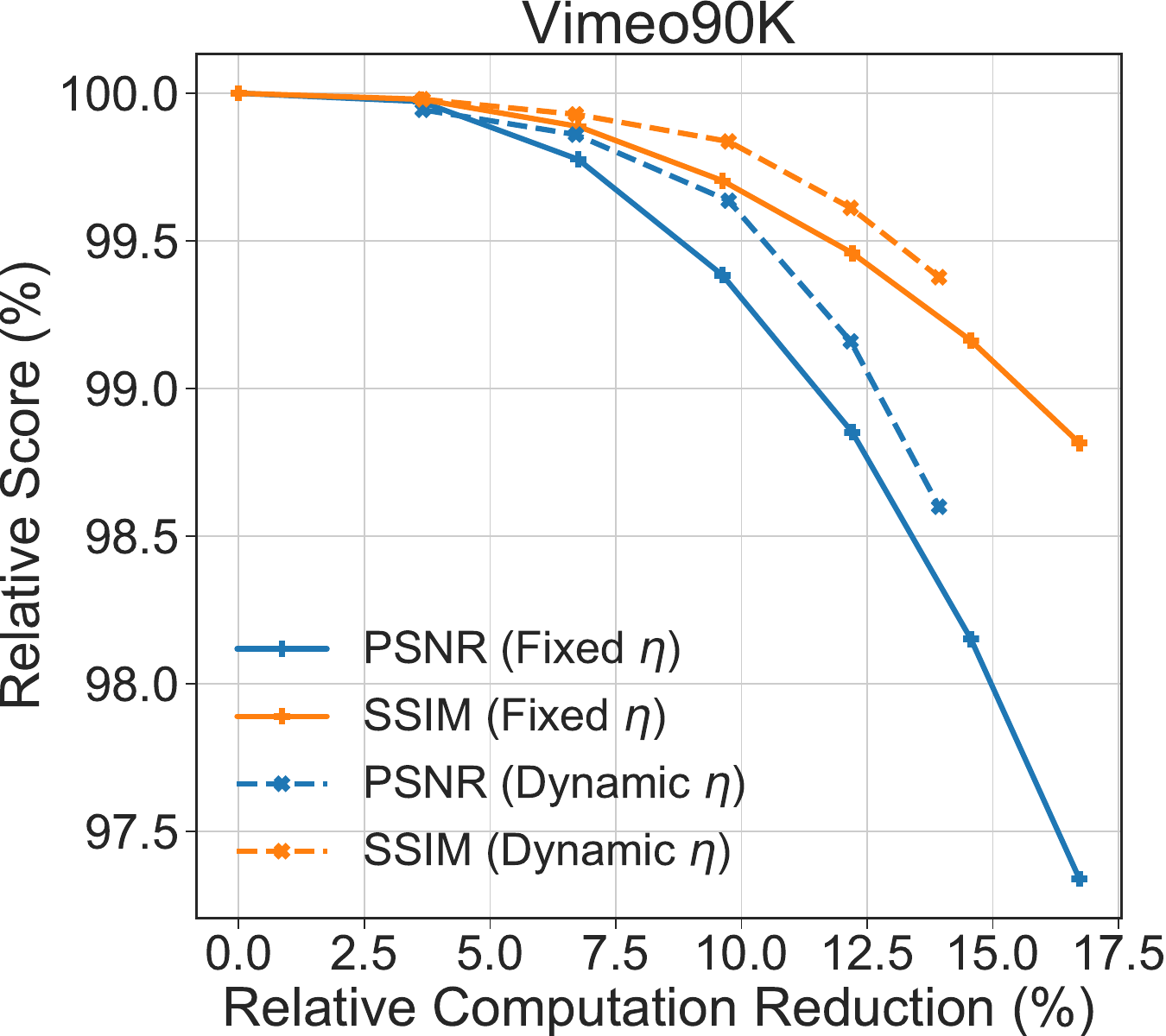}
	\includegraphics[width=0.245\linewidth]{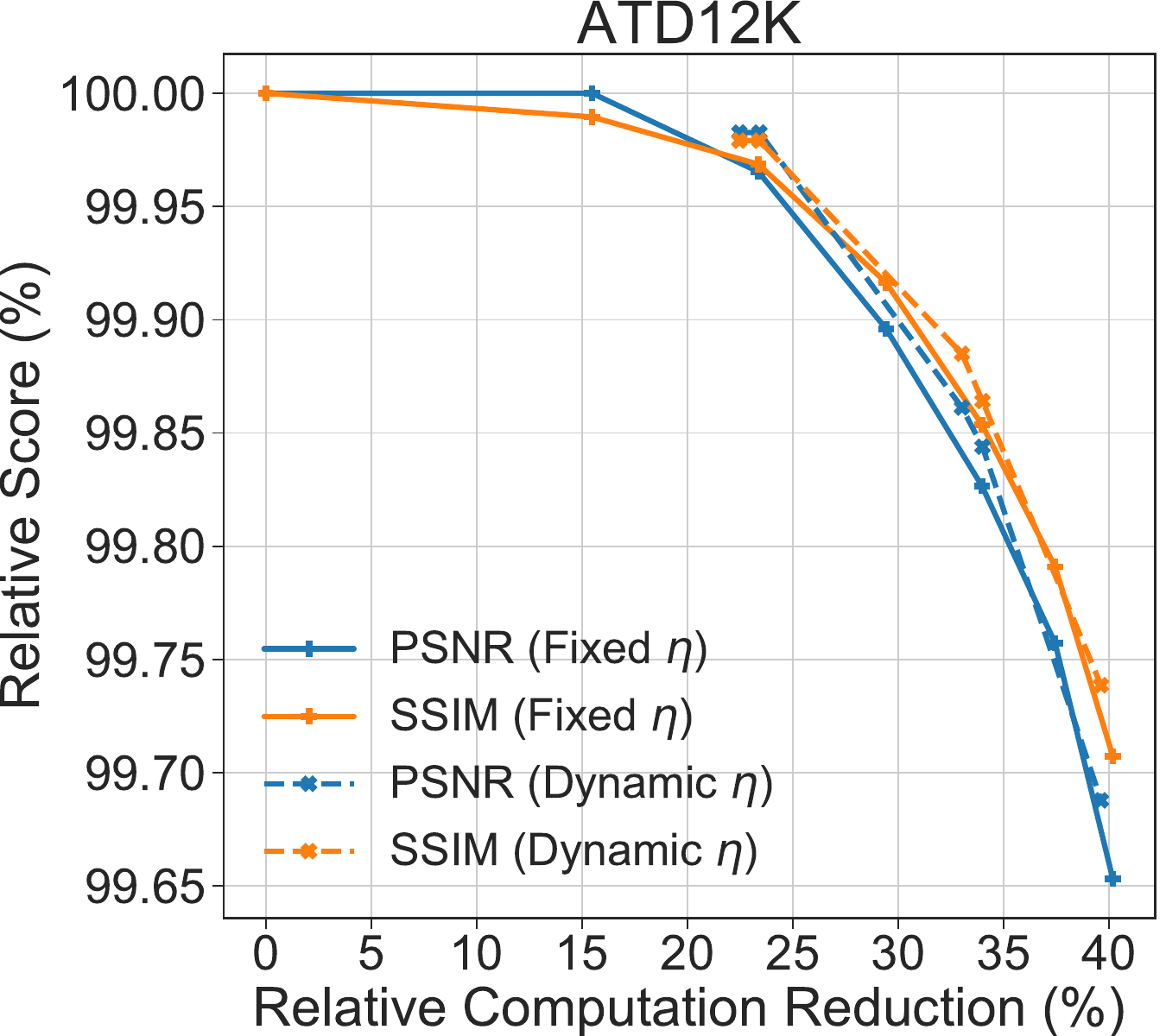}
	\includegraphics[width=0.245\linewidth]{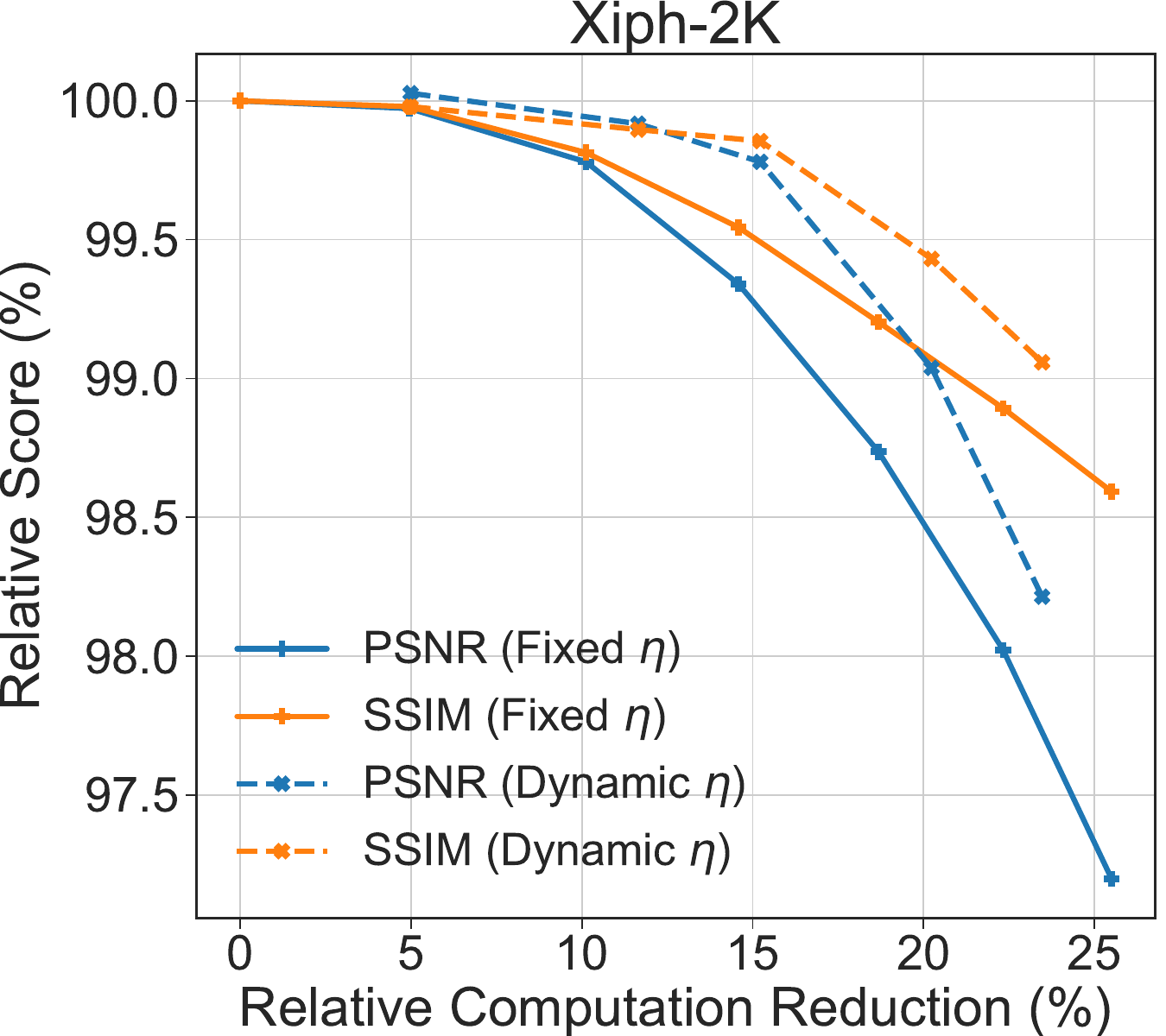}
	\includegraphics[width=0.245\linewidth]{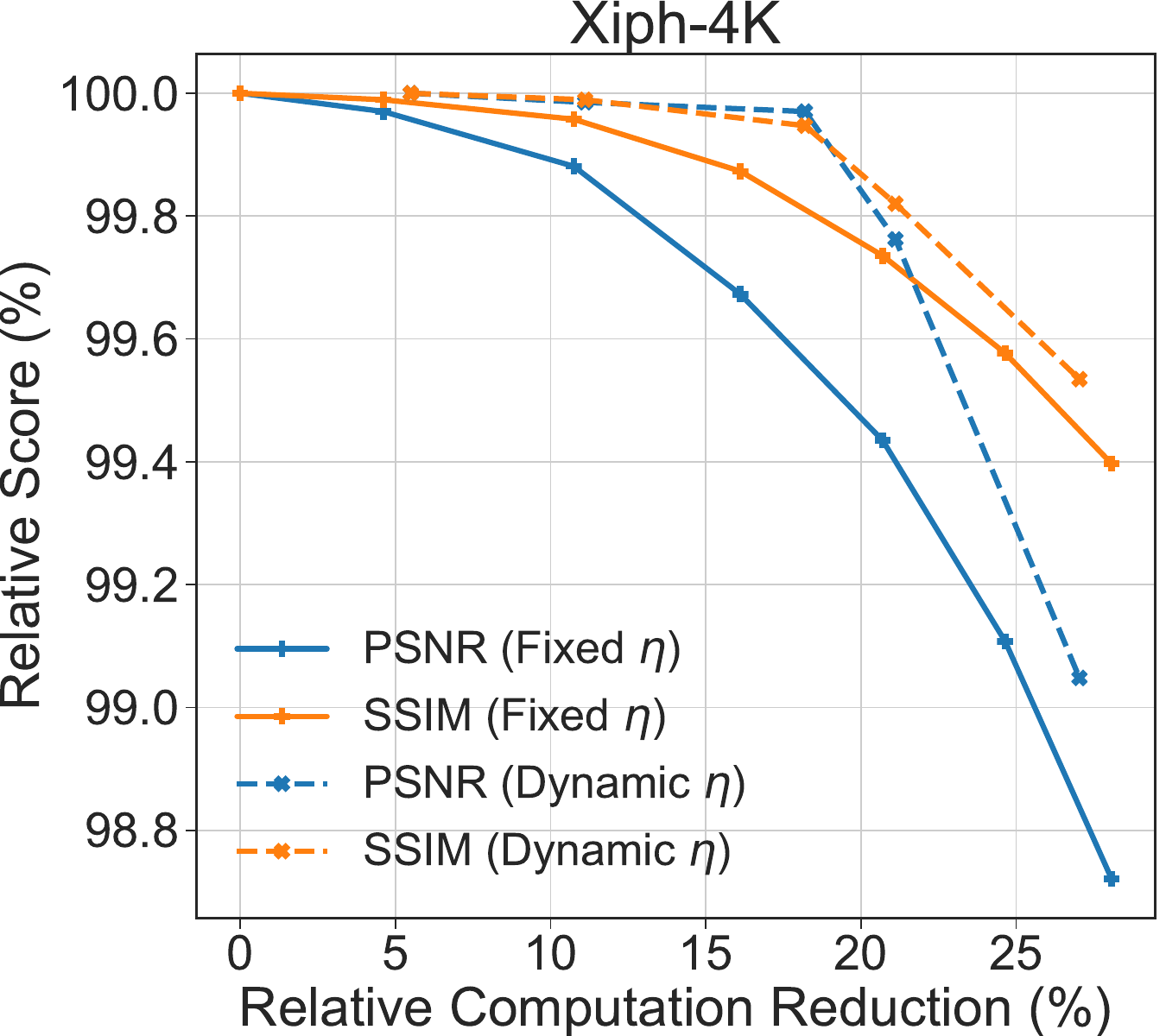}
	\caption{Analysis of accuracy vs efficiency under different compression threshold ratio selection approaches on multiple datasets.}
	\label{fig:12}
\end{figure*}

To verify proposed dynamic compression threshold ratio selection approach for instance-aware efficient frame interpolation, we vary computation cost regularization parameter $\beta$ during the second training step and record results of these dynamic models on diverse datasets, which are presented in Table~\ref{tab:6}. In this ablation, we set $m=4$ and threshold ratio candidates as 0.000, 0.005, 0.010 and 0.015 corresponding to above fixed threshold ratio experiments. The baseline method that does not use any wavelet compression approach is also listed in the first line for reference. In Table~\ref{tab:6}, the dynamic model tends to employ more computation to achieve better accuracy when given relatively small $\beta$, which is realized by selecting relatively small $\eta$ by the threshold classifier during end-to-end optimization. For a more intuitive comparison between fixed $\eta$ and dynamic $\eta$ settings, we further plot their accuracy vs efficiency trade-off line charts in Fig.~\ref{fig:12}. As is depicted, under the same computation cost, the relative improvement of dynamic selection approach against fixed threshold method behaves different in regard to both $\beta$ and datasets.

\begin{table*}[t]
	\centering
	\renewcommand{\arraystretch}{1.0}
	{\scriptsize
		\setlength\tabcolsep{4.8pt}
		\caption{Ablation study of computation cost regularization parameter $\beta$ for accuracy vs efficiency trade-off on multiple datasets.}
		\resizebox{0.98\textwidth}{!}{
			\begin{tabular}{cccccccccccccc}
				\toprule
				\multirow{2}[1]{*}{Dynamic} & \multirow{2}[1]{*}{$\beta$} & \multicolumn{3}{c}{Vimeo90K} & \multicolumn{3}{c}{ATD12K} & \multicolumn{3}{c}{Xiph-2K} & \multicolumn{3}{c}{Xiph-4K} \\
				\cmidrule(lr){3-5} \cmidrule(lr){6-8} \cmidrule(lr){9-11} \cmidrule(lr){12-14}
				& & TFLOPs & PSNR & SSIM & TFLOPs & PSNR & SSIM & TFLOPs & PSNR & SSIM & TFLOPs & PSNR & SSIM \\
				\midrule
				\xmark & 0.0 & 0.090 & 35.71 & 0.9791 & 0.409 & 28.83 & 0.9566 & 1.746 & 36.40 & 0.9664 & 1.746 & 33.62 & 0.9453 \\
				\midrule
				\cmark & 0.1 & 0.087 & 35.69 & 0.9789 & 0.317 & 28.83 & 0.9564 & 1.659 & 36.41 & 0.9662 & 1.650 & 33.62 & 0.9453 \\
				\cmark & 0.5 & 0.084 & 35.66 & 0.9784 & 0.313 & 28.83 & 0.9564 & 1.543 & 36.37 & 0.9654 & 1.552 & 33.62 & 0.9452 \\
				\cmark & 1.0 & 0.081 & 35.58 & 0.9775 & 0.274 & 28.79 & 0.9555 & 1.480 & 36.32 & 0.9650 & 1.428 & 33.61 & 0.9448 \\
				\cmark & 3.0 & 0.079 & 35.41 & 0.9753 & 0.270 & 28.78 & 0.9553 & 1.393 & 36.05 & 0.9609 & 1.377 & 33.54 & 0.9436 \\
				\cmark & 5.0 & 0.077 & 35.21 & 0.9730 & 0.247 & 28.74 & 0.9541 & 1.336 & 35.75 & 0.9573 & 1.274 & 33.30 & 0.9409 \\
				\bottomrule
		\end{tabular}}
		\label{tab:6}}
\end{table*}

\begin{figure}[t]
	\includegraphics[width=0.98\linewidth]{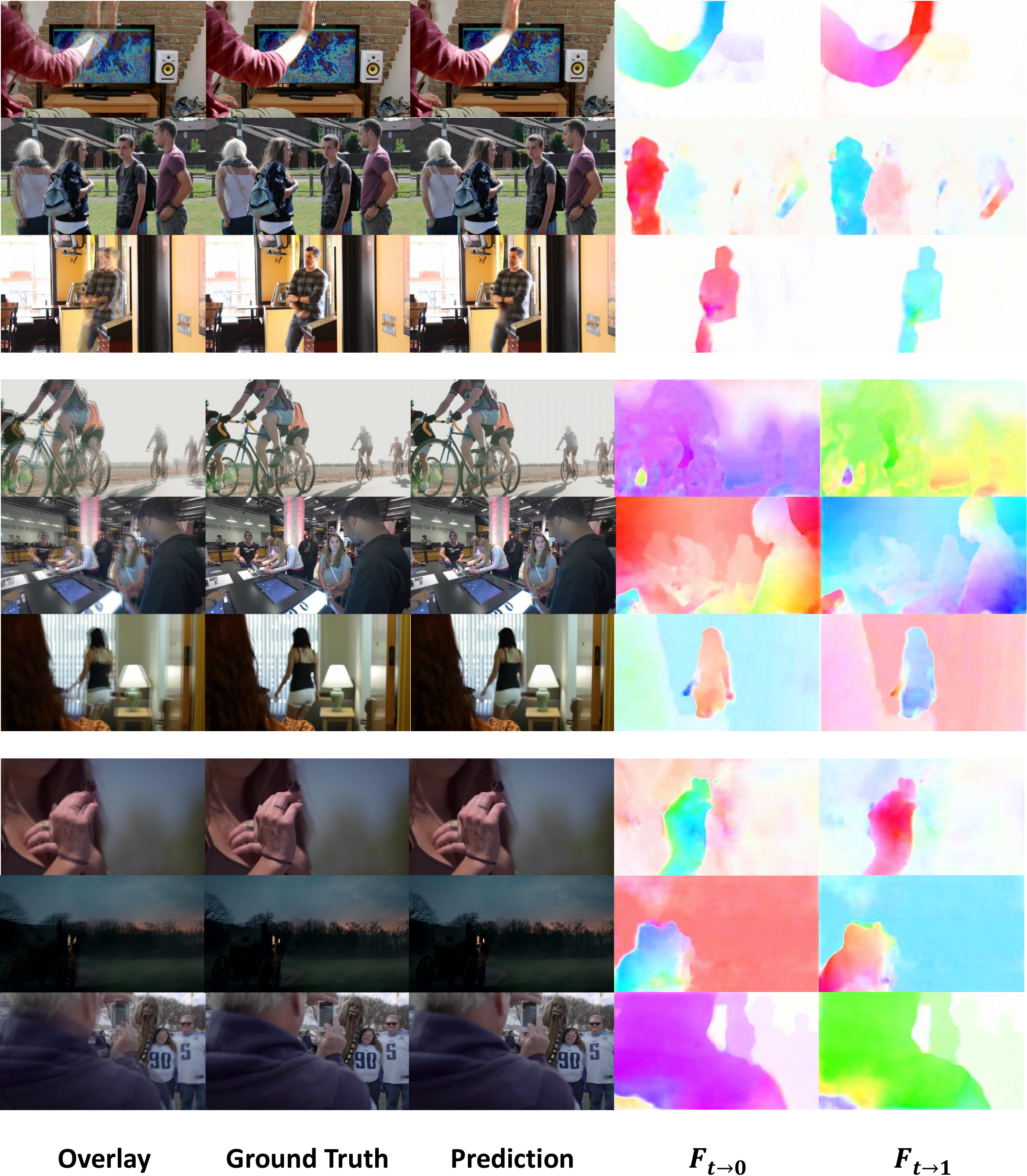}
	\caption{Visualization results of WaveletVFI for selecting different compression threshold ratios. Top, middle and bottom groups stand for selecting threshold ratio $\eta$ as 0.005, 0.010 and 0.015 respectively.}
	\label{fig:13}
\end{figure}

For the first factor, the relative improvement is more obvious when $\beta$ is set to a medium value. It is because that in this case the threshold classifier can have more chance to select different threshold candidates by learning the instance difference for frame interpolation. On the other hand, dynamic method tends to degrade to fixed method when given relatively large or small $\beta$. Therefore, we employ $\beta=1$ in WaveletVFI due to its largest performance gain. As for the second factor, the relative improvement is more obvious when the variation of spatial-temporal texture distribution in this dataset is larger. To prove it, we use FastFlowNet~\cite{Kong_2021_ICRA} to estimate inter-frame optical flow and analyze their motion magnitude statistics, whose results are shown in Table~\ref{tab:5} and Fig.~\ref{fig:11}. As can be seen in Fig.~\ref{fig:12}, there is almost no difference on ATD12K since it has relatively large motion variation but relatively small texture variation. On the other hand, the relative computation reduction under the same accuracy is much more obvious on the challenging Xiph-4K dataset. It is because that both the texture and the motion distribution variation on Xiph-4K are relatively large. In summary, proposed dynamic threshold selection approach in wavelet domain can achieve similar accuracy as the baseline method, while reducing computation cost by 10.0\%, 33.0\%, 15.2\% and 18.2\% on Vimeo90K, ATD12K, Xiph-2K and Xiph-4K benchmarks respectively.

\subsection{Discussion on Dynamic Threshold Selection}
In this part, we analyze the dynamic prediction results of the compression threshold ratio classifier on Vimeo90K test set with $\beta$ set to 1 during the second training step. Prediction results with different threshold ratio selection are visualized in Fig.~\ref{fig:13}. Intuitively, we can get following conclusions. \textbf{1)} In the top group of Fig.~\ref{fig:13}, when input frames come from rich and clear texture, and the inter-frame motion is relatively simple, the threshold classifier tends to estimate a small $\eta$, that consumes more computation to synthesize more reliable high-frequency texture of the target frame. \textbf{2)} In the middle group of Fig.~\ref{fig:13}, when input frames come from rich and clear texture, and the inter-frame motion is relatively complex, the threshold classifier tends to estimate a medium $\eta$, that reduces some computation for $\mathcal{N}_{\mathrm{WS}}$ to remove some unreliable high-frequency texture of the target frame. \textbf{3)} In the bottom group of Fig.~\ref{fig:13}, when input frames come from simple and blurry texture, and the inter-frame motion is relatively complex, the threshold classifier tends to select a large $\eta$, that consumes less computation to synthesize less unreliable high-frequency texture of the target frame.

\subsection{Limitations}
Currently, our framework only predicts the single middle frame, where $t=0.5$. For interpolating multiple intermediate frames, it can work in a recursive manner, but which may lead to error accumulation. This problem can be solved to some extent by modeling multiple discrete intermediate optical flow with a temporal encoding conditional input like IFRNet~\cite{Kong_2022_CVPR}, that can approximate arbitrary time interpolation.

\section{Conclusion}
To our best knowledge, it is the first time that the spatial redundancy problem in frame interpolation is studied in detail, which is particularly important with the popularity of high-resolution displays. In this work, we have proposed a novel frame interpolation algorithm in wavelet domain to achieve on par accuracy with SOTA methods but with better efficiency. Our method exploits the sparse representation in wavelet decomposition and employs sparse convolution to predict multi-scale wavelet coefficients in certain critical areas for computation reduction. Moreover, we have proposed a dynamic threshold selection approach to better allocate computation for each input sample. Experiments on the traditional low resolution, current high resolution and animation frame interpolation benchmarks demonstrate the effectiveness of proposed contributions, which can significantly reduce the overall resource consumption while maintaining advanced VFI accuracy. Since our approaches are orthogonal and complements other efficient methods, such as channel pruning, we hope proposed WaveletVFI can benefit the related communities.

\bibliographystyle{IEEEtran}
\bibliography{reference}

\vfill

\end{document}